%% file: main.tex
\documentclass[10pt,twocolumn,letterpaper]{article}

\usepackage{iccv}
\usepackage{times}
\usepackage{graphicx}
\usepackage{amsmath}
\usepackage{amssymb}
\usepackage[dvipsnames]{xcolor}

\usepackage{subcaption}
\usepackage{booktabs}
\usepackage{multirow}
\usepackage{float}
\usepackage{colortbl} %

\usepackage{pifont}  %
\usepackage{cite}  %
\usepackage{bm}  %
\usepackage{bbm} %
\usepackage{mathtools}  %
\usepackage[font=small,labelfont=bf]{caption}  %
\usepackage{blindtext}  %

\usepackage[pagebackref=true,breaklinks=true,letterpaper=true,colorlinks,bookmarks=false]{hyperref}

\iccvfinalcopy %

\def\iccvPaperID{4102} %

\input{sections/9_custom_commands}

\ificcvfinal\fi

\begin{document}

\title{Relational Embedding for Few-Shot Classification}
\author{Dahyun Kang \quad Heeseung Kwon \quad Juhong Min \quad Minsu Cho\vspace{0.15cm}\\ 
Pohang University of Science and Technology (POSTECH), South Korea\\
{\small \url{http://cvlab.postech.ac.kr/research/RENet}}
}

\maketitle
\ificcvfinal\thispagestyle{empty}\fi

\input{sections/0_abstract}

\input{sections/1_intro}
\input{sections/2_related_work}
\input{sections/3_preliminary}
\input{sections/4_method}
\input{sections/5_experiments}
\input{sections/6_conclusion}
\input{sections/7_acknowledgements}

{\small
\bibliographystyle{ieee_fullname}
\bibliography{egbib}
}

\clearpage
\renewcommand{\theequation}{a.\arabic{equation}}
\renewcommand{\thetable}{a.\arabic{table}}
\renewcommand{\thefigure}{a.\arabic{figure}}
\renewcommand*{\thefootnote}{\arabic{footnote}}
\renewcommand\thesection{\Alph{section}}
\setcounter{section}{0}
\input{sections/8_supp}

\end{document}

%% file: sections/9_custom_commands.tex
\newcommand{\itmini}{\textit{mini}}
\newcommand{\ittiered}{\textit{tiered}}
\newcommand{\Real}{\mathbb{R}}

\newcommand{\bs}{\mathbf{s}}
\newcommand{\bC}{\mathbf{C}}

\newcommand{\bq}{\mathbf{q}}
\newcommand{\bp}{\mathbf{p}}
\newcommand{\bF}{\mathbf{F}}
\newcommand{\bA}{\mathbf{A}}
\newcommand{\bI}{\mathbf{I}}
\newcommand{\bx}{\mathbf{x}}
\newcommand{\bZ}{\mathbf{Z}}

\newcommand{\bR}{\mathbf{R}}
\newcommand{\bt}{\mathbf{t}}

\newcommand{\bw}{\mathbf{w}}
\newcommand{\bz}{\mathbf{z}}
\newcommand{\bb}{\mathbf{b}}

\newcommand{\texts}{{\text{s}}}
\newcommand{\textq}{{\text{q}}}

\newcommand{\ours}{RENet\xspace}
\newcommand{\abbself}{SCR\xspace}
\newcommand{\abbcross}{CCA\xspace}

\colorlet{orange}{green!10!orange!90!}

\newcommand{\red}[1]{\textcolor{red}{#1}}

\newcommand{\smallbreakparagraph}[1]{\smallbreak \noindent \textbf{#1}}

\newcommand{\cmark}{\ding{51}}
\newcommand{\xmark}{\ding{55}}

\definecolor{grey}{rgb}{0.9, 0.9, 0.9}
\newcommand{\ccol}{\cellcolor{grey}}

\usepackage{xr}
\makeatletter
\newcommand*{\addFileDependency}[1]{%
  \typeout{(#1)}
  \@addtofilelist{#1}
  \IfFileExists{#1}{}{\typeout{No file #1.}}
}
\makeatother

%% file: sections/0_abstract.tex
\begin{abstract}
We propose to address the problem of few-shot classification by meta-learning ``what to observe'' and ``where to attend'' in a relational perspective.
Our method leverages relational patterns within and between images via self-correlational representation (SCR) and cross-correlational attention (CCA).
Within each image, the SCR module transforms a base feature map into a self-correlation tensor and learns to extract structural patterns from the tensor. 
Between the images, the CCA module computes cross-correlation between two image representations and learns to produce co-attention between them.
Our Relational Embedding Network (\ours) combines the two relational modules to learn relational embedding in an end-to-end manner.
In experimental evaluation, it achieves consistent improvements over state-of-the-art methods on four widely used few-shot classification benchmarks of \itmini ImageNet, \ittiered ImageNet, CUB-200-2011, and CIFAR-FS. 
\end{abstract}

%% file: sections/1_intro.tex
\section{Introduction}
Few-shot image classification~\cite{fei2006one, matchingnet, ravi2016optimization, koch2015siamese} aims to learn new visual concepts from a small number of examples.
The task is defined to classify a given query image into target classes, each of which is unseen during training and represented by only a few support images.
Recent methods~\cite{matchingnet, protonet, tadam, allen2019infinite, li2019revisiting, tewam, can, feat, deepemd, conceptlearners} tackle the problem by meta-learning a deep embedding function such that the distance between images on the embedding space conforms to their semantic distance. 
The learned embedding function, however, often overfits to irrelevant features~\cite{geirhos2018imagenet, brendel2019approximating, crosstransformers} and thus fails to transfer to new classes not yet observed in training.
While deep neural features provide rich semantic information, it remains challenging to learn a generalizable embedding without being distracted by spurious features.

\begin{figure}[t!]
	\centering
	\small
    \includegraphics[width=\linewidth]{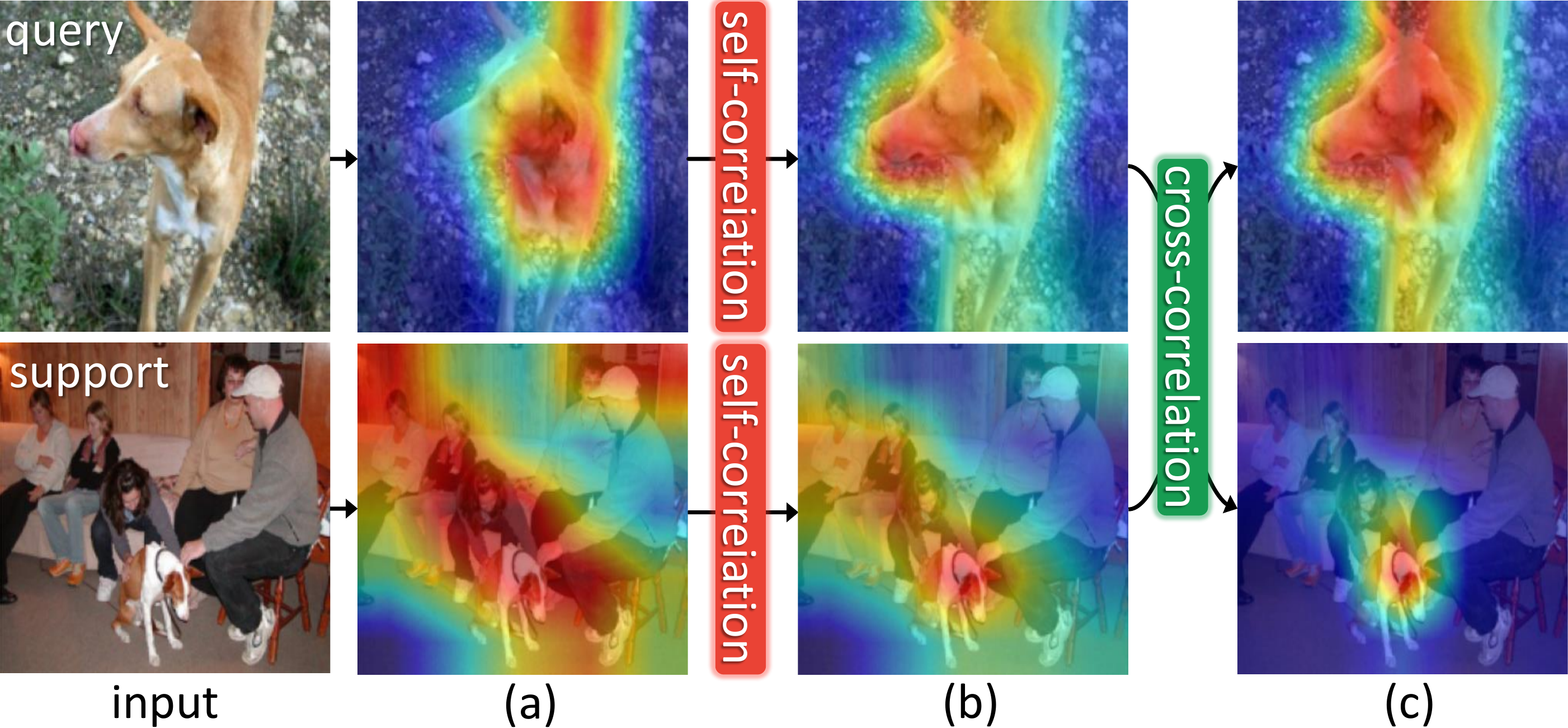}
	\caption{
    \textbf{Relational embedding process and its attentional effects.} 
    The base image features are transformed via self-correlation to capture structural patterns within each image and then co-attended via cross-correlation to focus on semantically relevant contents between the images.
    (a), (b), and (c) visualize the activation maps of base features, self-correlational representation, and cross-correlational attention, respectively. See Sec.~\ref{sec:qual} for details. 
	}
\label{fig:teaser}
\vspace{-4mm}
\end{figure}

The central tenet of our approach is that relational patterns, \ie, meta-patterns, may generalize better than individual patterns; an item obtains a meaning in comparison with  other items in a system, and thus relevant information can be extracted from the relational structure of items. On this basis, we propose to learn ``what to observe'' and ``where to attend'' in a relational perspective and combine them to produce relational embeddings for few-shot learning.

We achieve this goal by leveraging relational patterns {\em within} and {\em between} images via (1) {\em self-correlational representation} (SCR) and (2) {\em cross-correlational attention} (CCA).
The SCR module transforms a base representation into its self-correlation tensor and learns to extract structural patterns from it. 
Self-correlation of a deep feature map encodes rich semantic structures by correlating each activation of the feature map to its neighborhood. We perform representation learning on top of it to make relevant structural patterns of the image stand out (Fig.~\ref{fig:teaser}~(a)$\rightarrow$(b)).
On the other hand, the CCA module computes cross-correlation between two image representations and learns to produce co-attention from it. 
Cross-correlation encodes semantic correspondence relations between the two images. 
We learn high-dimensional convolutions on the cross-correlation tensor to refine it via convolutional matching and produce adaptive co-attention based on semantic relations between the query and the support (Fig.~\ref{fig:teaser}~(b)$\rightarrow$(c)).

The proposed method combines the two modules to learn relational embeddings in an end-to-end manner; it extracts relational patterns {\em within} each image (via SCR), generates relational attention {\em between} the images (via CCA), and aggregates the cross-attended self-correlation representations to produce the embeddings for few-shot classification.
Experiments on four standard benchmark datasets demonstrate that the proposed SCR and CCA modules are effective at highlighting the target object regions and significantly improve few-shot image classification accuracy.

\begin{figure*}[t!]
	\centering
	\small
    \includegraphics[width=1\textwidth]{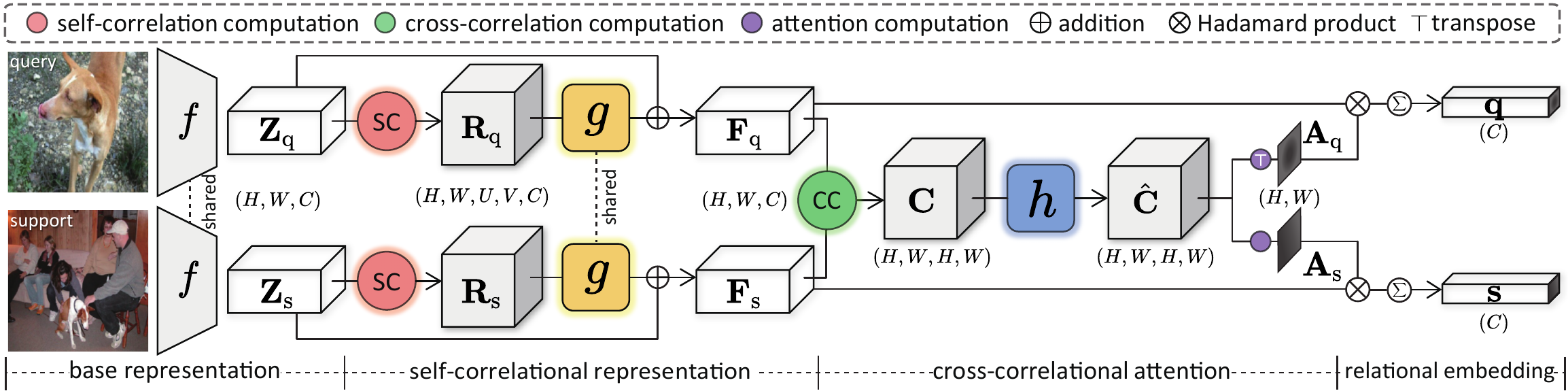}
	\caption{
	\textbf{Overall architecture of \ours.}
    The base representations, $\bZ_\textq$ and $\bZ_\texts$, are transformed to self-correlation tensors, $\bR_\textq$ and $\bR_\texts$, which are then updated by the convolutional block $g$ to self-correlational representations, $\bF_\textq$ and $\bF_\texts$, respectively.
    The cross-correlation $\bC$ is computed between the pair of image representations and then refined by the convolutional block $h$ to $\hat\bC$, which is bidirectionally aggregated to generate co-attention maps, $\bA_\textq$ and $\bA_\texts$. 
    These co-attention maps are applied to corresponding image representations, $\bF_\textq$ and $\bF_\texts$, and their attended features are aggregated to produce the final relational embeddings, $\bq$ and $\bs$, respectively. 
    }
\label{fig:architecture}
\end{figure*}

%% file: sections/2_related_work.tex
\section{Related work}
\smallbreakparagraph{Few-shot classification.}
Recent few-shot classification methods are roughly categorized into three approaches.
The metric-based approach aims to learn an embedding function that maps images to a metric space such that the relevance between a pair of images is distinguished based on their distance~\cite{koch2015siamese, matchingnet, protonet, tadam, allen2019infinite, li2019revisiting, ctm, tewam, can, feat, deepemd, lifchitz2021local, crosstransformers}.
The optimization-based approach meta-learns how to rapidly update models online given a small number of support samples~\cite{maml, bmaml, leo, ravi2016optimization, mtl}.
The two aforementioned lines of work formulate few-shot classification as a meta-learning problem~\cite{schmidhuber1987evolutionary, bengio1990learning, hochreiter2001learning}.
The transfer-learning approach~\cite{closer, rfs, dhillon2019baseline, wang2020few, negmargin, gidaris2018dynamic, qi2018low, rodriguez2020embedding, ziko2020laplacian} has recently shown that the standard transfer learning procedure~\cite{yosinski2014transferable, oquab2014} of early pre-training and subsequent fine-tuning is a strong baseline for few-shot learning with deep backbone networks.
Among these, our work belongs to the metric-based approach.
The main idea behind a metric-based few-shot classifier is that real images are distributed on some manifolds of interest, thus the embedding function adequately trained on the training classes can be transferred to embed images of unseen target classes by interpolating or extrapolating the features~\cite{tenenbaum1998mapping, facenet}.
Our work improves the transferability of embedding by learning self- and cross-relational patterns that can better generalize to unseen classes.

\smallbreakparagraph{Self-correlation.}
Self-correlation or self-similarity reveals a structural layout of an image by measuring similarities of a local patch within its neighborhood~\cite{shechtman2007matching}.
Early work uses the self-correlation itself as a robust descriptor for visual correspondence~\cite{torabi2013local}, object detection~\cite{deselaers2010global}, and action recognition~\cite{junejo2011view,junejo2008cross}. 
Recent work of~\cite{fcss, zheng2021spatially, stss} adopts self-correlation as an intermediate feature transform for a deep neural network and shows that it helps the network learn an effective representation for semantic correspondence~\cite{fcss}, image translation~\cite{zheng2021spatially}, and video understanding~\cite{stss}.  
Inspired by the work, we introduce the SCR module for few-shot classification. 
Unlike self-correlation used in the previous work, however, our SCR module uses channel-wise self-correlation to preserve rich semantic information for image recognition.
Note that while self-attention~\cite{nlsa,lsa} also computes self-correlation values as attention weights for aggregation, it does not use the self-correlation tensor directly for representation learning and thus is distinct from this line of research.

\smallbreakparagraph{Cross-correlation.}
Cross-correlation has long been used as a core component for a wide range of correspondence-related problems in computer vision. 
It is commonly implemented as a cost-volume or correlation layer in a neural network, which computes matching costs or similarities between two feature maps, and is used for stereo-matching~\cite{zbontar2016stereo,luo2016efficient}, optical flow~\cite{dosovitskiy2015flownet,sun2018pwc,vcn}, visual correspondence~\cite{hpf, dhpf, ncnet, ancnet, chm}, semantic segmentation~\cite{sun2020mining, hsnet}, video action recognition~\cite{wang2020video, motionsqueeze}, video object segmentation~\cite{oh2019video, hu2018videomatch}, among others. 
Some recent few-shot classification methods~\cite{deepemd, li2019revisiting, can, crosstransformers} adopt cross-correlation between a query and each support to identify relevant regions for classification. 
However, none of them~\cite{deepemd, li2019revisiting, crosstransformers, can} leverage geometric relations of features in cross-correlation and they often suffer from unreliable correlation due to the large variation of appearance.
Unlike these previous methods, our \abbcross module learns to refine the cross-correlation tensor with 4D convolution, filtering out geometrically inconsistent correlations~\cite{ncnet, chm}, to obtain reliable co-attention. 
In our experiment, we provide an in-depth comparison with the most related work of~\cite{can}.

Our contribution can be summarized as follows:
\begin{itemize}
    \item We propose to learn the self-correlational representation for few-shot classification, which extracts transferable structural patterns within an image.
    \item We present the cross-correlational attention module for few-shot classification, which learns reliable co-attention between images via convolutional filtering.    
    \item Experiments on four standard benchmarks show our method achieves the state of the art, and ablation studies validate the effectiveness of the components.
\end{itemize}

%% file: sections/3_preliminary.tex
\section{Preliminary on few-shot classification}
Few-shot classification aims to classify images into target classes given only a few images for each class. 
Deep neural networks are vulnerable to overfitting with such a small amount of annotated samples, and most few-shot classification methods~\cite{matchingnet, ravi2016optimization, protonet} thus adopt a meta-learning framework with episodic training for few-shot adaptation. 
In few-shot classification, a model is optimized using training data $\mathcal{D}_{\text{train}}$ from classes $\mathcal{C}_{\text{train}}$ and then evaluated on test data $\mathcal{D}_{\text{test}}$ from unseen classes $\mathcal{C}_{\text{test}}$ where $\mathcal{C}_{\text{train}} \cap \mathcal{C}_{\text{test}} = \varnothing$.
Both $\mathcal{D}_{\text{train}}$ and $\mathcal{D}_{\text{test}}$ consist of multiple episodes, each of which contains a query set $\mathcal{Q}=(\bI_\textq, y_\textq)$ and a support set $\mathcal{S}=\{(\bI_\texts^{(l)}, y_\texts^{(l)})\}_{l=1}^{NK}$ of $K$ image-label pairs for each $N$ classes, \ie, $N$-way $K$-shot episode~\cite{fei2006one, matchingnet}.
During training, we iteratively sample an episode from $\mathcal{D}_{\text{train}}$ and train the model to learn a mapping from $(\mathcal{S}, \bI_\textq)$ to $y_\textq$.
During testing, the model uses the learned mapping to classify $\bI_\textq$ as one of $N$ classes in the support set $\mathcal{S}$ sampled from $\mathcal{D}_{\text{test}}$.

%% file: sections/4_method.tex
\section {Our approach}
In this section, we introduce the \textit{\textbf{R}elational \textbf{E}mbedding \textbf{Net}work} (\textbf{\ours}) that addresses the challenge of generalization to unseen target classes in a relational perspective.
Figure~\ref{fig:architecture} illustrates the overall architecture, consisting of two main learnable modules: self-correlational representation (\abbself) module and cross-correlational attention (\abbcross) module.
We first present a brief overview of the proposed architecture in Sec.~\ref{sec_overview}.
We then present technical details of \abbself and \abbcross in Sec.~\ref{sec_SCR} and Sec.~\ref{sec_CCA} respectively, and describe our training objective in Sec.~\ref{sec_CORE}.

\subsection{Architecture overview}
\label{sec_overview}
Given a pair of a query and one of support images, $\bI_\textq$ and $\bI_\texts$, a backbone feature extractor provides base representations, $\bZ_\textq$ and $\bZ_\texts \in \mathbb{R}^{H \times W \times C}$.
The SCR module transforms the base representations to self-correlational representations, $\bF_\textq$ and $\bF_\texts \in \mathbb{R}^{H \times W \times C}$, by analyzing feature correlations within an image representation in a convolutional manner.
The CCA module then takes the self-correlational representations to generate co-attention maps, $\bA_\textq$ and $ \bA_\texts \in \Real^{H \times W}$, which give spatial attention weights on aggregating $\bF_\textq$ and $\bF_\texts$ to image embeddings, $\bq$ and $\bs \in \Real^{C}$.
This process illustrated in Fig.~\ref{fig:architecture} is applied to all support images $\bI_\texts \in \mathcal{S}$ in parallel, and then the query is classified as the class of its nearest support embedding.

\subsection{Self-Correlational Representation (\abbself)}~\label{sec_SCR}
\begin{figure}[t!]
\begin{subfigure}[t]{.49\linewidth}
\includegraphics[width=1\linewidth]{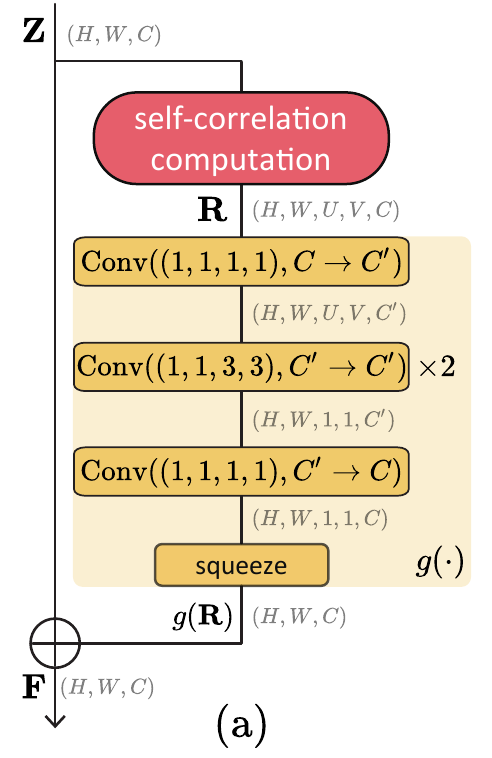}
\captionlistentry{}
\label{fig:self_architecture}
\end{subfigure}
\hfill
\begin{subfigure}[t]{.49\linewidth}
\includegraphics[width=1\linewidth]{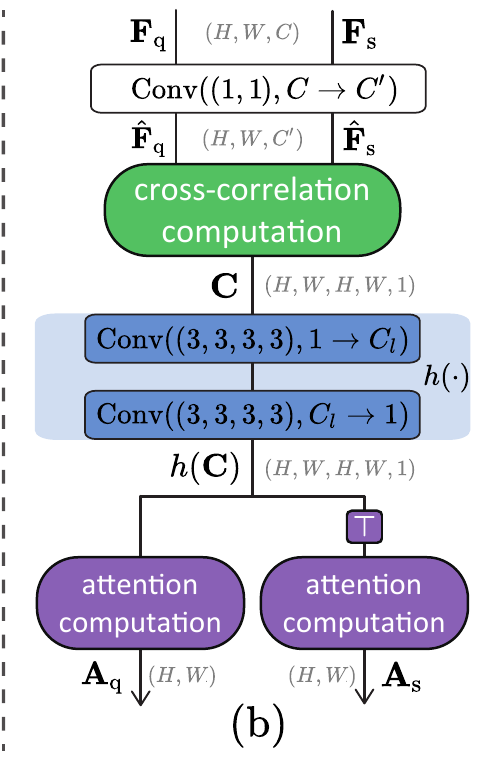}
\captionlistentry{}
\label{fig:cross_architecture}
\end{subfigure}
\vspace{-3mm}
\caption{
\textbf{Architecture of \abbself and \abbcross modules.}
\textbf{(a)}: The \abbself module captures relational patterns in the input self-correlation $\bR$ by convolving it over $U \times V$ dimensions.
The result $g(\bR)$ is added to the base representation $\bZ$ to form the self-correlational representation $\bF$ (Eq.~\ref{eq:ss}).
\textbf{(b)}: The \abbcross module refines the cross-correlation which will be summarized into co-attention maps, $\bA_\textq$ and $\bA_\texts$ (Eq.~\ref{eq:attn}).
\label{fig:self_cross_architecture}
}
\vspace{-2mm}
\end{figure}

The \abbself module takes the base representation $\bZ$\footnote{For notational simplicity, we omit subscripts $\textq$ and $\texts$ in this subsection.} and transforms it to focus more on relevant regions in an image, preparing a reliable input to the \abbcross module that analyzes feature correlations between a pair of different images.
Figure~\ref{fig:self_architecture} illustrates the architecture of the \abbself module.

\paragraph{Self-correlation computation.}
Given a base representation $\bZ \in \mathbb{R}^{H \times W \times C}$, we compute the Hadamard product of a $C$-dimensional vector at each position $\bx \in [1,H] \times [1,W]$ and those at its neighborhood and collect them into a self-correlation tensor $\bR \in \Real^{H \times W \times U \times V \times C}$. With an abuse of notation, the tensor $\bR$ can be represented as a function with a $C$-dimensional vector output:
\begin{align}
    \bR(\bx, \bp) &= \frac{\bZ(\bx)}{\lVert \bZ(\bx) \rVert} \odot \frac{\bZ(\bx + \bp)}{\lVert \bZ(\bx + \bp) \rVert}, 
    \label{eq:self_computation}
\end{align}
where $\bp \in [-d_U, d_U] \times [-d_V, d_V]$ corresponds to a relative position in the neighborhood window such that $2d_U + 1 = U$ and $2d_V + 1 = V$, including the center position.
Note that the edges of the feature map are zero-padded for sampling off the edges.
The similar type of self-correlation, \ie, self-similarity, has been used as a relational descriptor for images and videos that suppresses variations in appearance and reveals structural patterns~\cite{shechtman2007matching}.
Unlike the previous methods~\cite{shechtman2007matching, deselaers2010global, junejo2008cross}, which reduce a pair of feature vectors into a scalar correlation value, we use the channel-wise correlation, preserving rich semantics of the feature vectors for classification.

\paragraph{Self-correlational representation learning.}
To analyze the self-correlation patterns in $\bR$, we apply a series of 2D convolutions along $U \times V$ dimensions.
As shown in Fig.~\ref{fig:self_architecture}, 
the convolutional block follows a bottleneck structure~\cite{inceptions} for computational efficiency, which is comprised of a point-wise convolution layer for channel size reduction, two $3\times3$ convolution layers for transformation, and another point-wise convolution for channel size recovery.
Between the convolutions, batch normalization\cite{batchnorm} and ReLU~\cite{relu} are inserted.
This convolutional block $g(\cdot)$ gradually aggregates local correlation patterns without padding, thus reducing their spatial dimensions from $U \times V$ to $1 \times 1$ such that the output $g(\bR)$ has the same size with $\bZ$, \ie, $g: \Real^{H \times W \times U \times V \times C} \rightarrow \Real^{H \times W \times C}$. 
This process of analyzing structural patterns may be complementary to appearance patterns in the base representation $\bZ$. 
We thus combine the two representations to produce the self-correlational representation $\bF \in \mathbb{R}^{H \times W \times C}$:
\begin{eqnarray}
	\bF = g(\bR) + \bZ,  
	\label{eq:ss}
\end{eqnarray}
which reinforces the base features with relational features and helps the few-shot learner better understand ``what to observe'' within an image.
Our experiments show that \abbself is robust to intra-class variations and helps generalization to unseen target classes.

\subsection{Cross-Correlational Attention (\abbcross)}~\label{sec_CCA}
The \abbcross module takes an input pair of query and support SCRs, $\bF_\textq$ and $\bF_\texts$, and produces corresponding attention maps, $\bA_\textq$ and $\bA_\texts$.
These spatial attention maps are used to aggregate each representation to an embedding vector.
Figure~\ref{fig:cross_architecture} visualizes the pipeline of the \abbcross module.

\smallbreakparagraph{Cross-correlation computation.}
We first transform both query and support representations, $\bF_{\textq}$ and $\bF_{\texts}\in \Real^{H \times W \times C}$, into more compact representations using a point-wise convolutional layer, reducing its channel dimension $C$ to $C'$.
From the outputs, $\hat{\bF}_{\textq}$ and $ \hat{\bF}_{\texts} \in \Real^{H \times W \times C'}$, we construct a 4-dimensional correlation tensor $\bC\in\Real^{H \times W \times H \times W}$:
\begin{eqnarray}
    \bC(\bx_\textq, \bx_\texts) = \mathrm{sim}(\hat{\bF}_\textq(\bx_\textq),\hat{\bF}_\texts(\bx_\texts)),
    \label{eq:cross_correlation}
\end{eqnarray}
where $\bx$ denotes a spatial position on the feature map and $\mathrm{sim}(\cdot, \cdot)$ means the cosine similarity between two features.

\smallbreakparagraph{Convolutional matching.}
The cross-correlation tensor $\bC$ may contain unreliable correlations, \ie, matching scores, due to the large appearance variations in the few-shot learning setup. 
To disambiguate those unreliable matches, we employ the convolutional matching process~\cite{ncnet,chm} that refines the tensor by 4D convolutions with  matching kernels; 4D convolution on the tensor plays the role of geometric matching by analyzing the consensus of neighboring matches in the 4D space.
As shown in Fig.~\ref{fig:cross_architecture}, the convolutional matching block $h(\cdot)$ consists of two 4D convolutional layers; the first convolution produces multiple correlation tensors with multiple matching kernels, increases channel size to $C_l$, and the second convolution aggregates them to a single 4D correlation tensor, \ie, $\hat{\bC} = h(\bC) \in \Real^{H\times W \times H \times W }$. 
Batch normalization and ReLU are inserted between the convolutions.
We empirically found that two 4D convolutional layers are sufficient for our \abbcross module.

\paragraph{Co-attention computation.}
From the refined tensor $\hat{\bC}$, we produce co-attention maps, $\bA_\textq$ and $\bA_\texts$, which reveal relevant contents between the query and the support.
The attention map for the query $\bA_\textq \in \Real^{H \times W}$ is computed by
\begin{eqnarray}
	\bA_\textq(\bx_\textq) = \frac{1}{H W}\sum_{\bx_\texts} \frac{\exp{(\hat{\bC}(\bx_\textq, \bx_\texts) / \gamma)} }{\sum_{\bx_\textq'}
	\exp{(\hat{\bC}(\bx_\textq', \bx_\texts) / \gamma)}},\label{eq:attn}
\end{eqnarray}
where $\bx$ is a position at the feature map and $\gamma$ is a temperature factor.
Since $\hat{\bC}(\bx_\textq, \bx_\texts)$ is a matching score between the positions $\bx_\textq$ and $\bx_\texts$, the attention value $\bA_\textq(\bx_\textq)$ of Eq.~(\ref{eq:attn}) can be interpreted as converting the matching score of  $\bx_\textq$, \ie, a position at the query image, to the average probability of $\bx_\textq$ being matched to a position at the support image.
The attention map for the support $\bA_\texts$ is similarly computed by switching the query and the support in Eq.~(\ref{eq:attn}).

These co-attention maps improve few-shot classification accuracy by meta-learning cross-correlational patterns and adapting ``where to attend'' with respect to the images given at test time.

\subsection{Learning relational embedding}~\label{sec_CORE}
In this subsection, we derive relational embeddings $\bq$ and $\bs \in \Real^C$ from $\bF_\textq, \bF_\texts, \bA_\textq$ and $\bA_\texts$.
We then conclude our method by describing the learning objective.

\paragraph{Attentive pooling.}
To obtain the final embedding of the query, $\bq \in \Real^{C}$, each position of $\bF_\textq \in \Real^{H \times W \times C}$ is multiplied by the spatial attention map $\bA_\textq \in \Real^{H \times W}$ followed by pooling:
\begin{eqnarray}
	\bq = \sum_{\bx_\textq} \bA_\textq(\bx_\textq) \bF_\textq(\bx_\textq).\label{eq:attentive_pooling_q}
\end{eqnarray}
Note that the elements of $\bA_\textq$ sum up to 1, and thus the attentive embedding $\bq$ is a convex combination of $\bF_\textq$ attended in the context of the support.
The final embedding of the support is computed similarly by attending the support feature map $\bF_\texts$ by $\bA_\texts$ followed by pooling:
\begin{eqnarray}
	\bs = \sum_{\bx_\texts} \bA_\texts(\bx_\texts) \bF_\texts(\bx_\texts).\label{eq:attentive_pooling_s}
\end{eqnarray}
On an $N$-way $K$-shot classification setting, this co-attentive pooling generates a set of $NK$ different views of a query, $\{\bq^{(l)} \}_{l = 1}^{NK}$, and a set of support embeddings attended in the context of the query, $\{\bs^{(l)} \}_{l = 1}^{NK}$.

\paragraph{Learning objective.}
\label{sec:learning_objective}
The proposed \ours is end-to-end trainable from scratch.
While most of the recent few-shot classification methods adopt the two-stage training scheme~\cite{leo, frn, feat, deepemd} of initial pre-training and subsequent episodic training, we adopt the single-stage training scheme~\cite{can, tadam} that jointly trains the proposed modules as well as the backbone network by combining two losses: the anchor-based classification loss $\mathcal{L}_{\text{anchor}}$ and the metric-based classification loss $\mathcal{L}_{\text{metric}}$.
First, $\mathcal{L}_{\text{anchor}}$ is computed with an additional fully-connected classification layer on top of average-pooled base representation $\bz_\textq$.
This loss guides the model to correctly classify a query of class $c \in \mathcal{C}_{\text{train}}$: 
\begin{eqnarray}
    \mathcal{L}_{\text{anchor}} = -\log \frac{\mathrm{exp}(\bw_{c}^{\top}\bz_\textq + \bb_{c})}{\sum_{c'=1}^{|\mathcal{C}_{\text{train}}|} \mathrm{exp}(\bw_{c'}^{\top}\bz_\textq + \bb_{c'})},
\label{eq:fc_classification_probability}
\end{eqnarray}
where $[\bw_{1}^{\top}, \cdots, \bw_{|\mathcal{C}_{\text{train}}|}^{\top}]$ and $[\bb_{1}, \cdots, \bb_{|\mathcal{C}_{\text{train}}|}]$ are weights and biases in the fully-connected layer, respectively.
Next, the metric-based loss $\mathcal{L}_{\text{metric}}$~\cite{matchingnet, protonet} is computed by cosine similarity between a query and support prototype embeddings.
Before computing the loss, we average the $K$ query embedding vectors each of which is attended in the context of $k^{\text{th}}$ support from $n^{\text{th}}$ class to compute $\{\bar{\bq}^{(n)}\}_{n=1}^{N}$.
Similarly, we average the $K$ support embeddings for each class to obtain a set of prototype embeddings: $\{\bar{\bs}^{(n)}\}_{n=1}^{N}$.
The metric-based loss guides the model to map a query embedding close to the prototype embedding of the same class:
\begin{eqnarray}
    \mathcal{L}_{\text{metric}} =-\log \frac{\mathrm{exp}(\mathrm{sim}(\mathbf{\bar{\bs}}^{(n)}, \mathbf{\bar{\bq}}^{(n)})/\tau)}{\sum_{n'=1}^{N}\mathrm{exp}(\mathrm{sim}(\mathbf{\bar{\bs}}^{(n')}, \mathbf{\bar{\bq}}^{(n')})/\tau)},
\label{eq:classification_probability}
\end{eqnarray}
where $\mathrm{sim}(\cdot, \cdot)$ is cosine similarity and $\tau$ is a scalar temperature factor.
At inference, the class of the query is predicted as that of the nearest prototype.

The objective combines the two losses: 
\begin{eqnarray}
    \mathcal{L} = \mathcal{L}_{\text{anchor}} + \lambda \mathcal{L}_{\text{metric}},\label{eq:losses}
\end{eqnarray}
where $\lambda$ is a hyper-parameter that balances the loss terms.
Note that the fully-connected layer involved in computing $\mathcal{L}_{\text{anchor}}$ is discarded during inference.

%% file: sections/5_experiments.tex
\section{Experimental results}
In this section, we evaluate \ours on standard benchmarks and compare the results with the recent state of the arts.
We also conduct ablation studies to validate the effect of the major components.
For additional results and analyses, we refer the readers to our appendix.

\subsection{Datasets}
For evaluation, we use four standard benchmarks for few-shot classification: \itmini ImageNet, \ittiered ImageNet, CUB-200-2011, and CIFAR-FS.
\textbf{\itmini ImageNet}~\cite{matchingnet} is a subset of ImageNet~\cite{russakovsky2015imagenet} consisting of 60,000 images uniformly distributed over 100 object classes.
The train/validation/test splits consist of 64/16/20 object classes, respectively.
\textbf{\ittiered ImageNet}~\cite{tieredimagenet} is a challenging dataset in which train/validation/test splits are disjoint in terms of super-classes from the ImageNet hierarchy, which typically demands better generalization than other datasets.
The respective train/validation/test splits consist of 20/6/8 super-classes, which are super-sets of 351/97/160 sub-classes.
\textbf{CUB-200-2011} (CUB)~\cite{cub} is a dataset for fine-grained classification of bird species, consisting of 100/50/50 object classes for train/validation/test splits, respectively.
Following the recent work of \cite{deepemd, feat}, we use pre-cropped images to human-annotated bounding boxes.
\textbf{CIFAR-FS}~\cite{cifarfs} is built upon CIFAR-100~\cite{cifar} dataset. 
Following the recent work of \cite{cifarfs}, we use the same train/validation/test splits consisting of 64/16/20 object classes, respectively.
For all the datasets, $\mathcal{D}_{\text{train}}$, $\mathcal{D}_{\text{val}}$, and $\mathcal{D}_{\text{test}}$ are disjoint in terms of object classes such that $\mathcal{C}_{\text{train}} \cap \mathcal{C}_{\text{val}} \, = \mathcal{C}_{\text{val}} \cap \mathcal{C}_{\text{test}} = \mathcal{C}_{\text{test}} \cap \mathcal{C}_{\text{train}} = \varnothing$.

\subsection{Implementation details}
We adopt ResNet12~\cite{resnet} following the recent few-shot classification work~\cite{shotfree, tadam, feat, deepemd}.
The backbone network takes an image with spatial size of $84 \times 84$ as an input and provides a base representation $\bZ \in \mathbb{R}^{5 \times 5 \times 640}$ followed by shifting its channel activations by the channel mean of an episode \cite{deepemd}.
For our CCA module, we adopt separable 4D convolutions~\cite{vcn} with kernel size of $3 \times 3 \times 3 \times 3$ for its effectiveness in approximating the original 4D convolutions~\cite{ncnet} as well as efficiency in terms of both memory and time.
The output of the 4D convolution $\hat{\bC}$ is normalized such that the entities in the pair of spatial map to be zero-mean and unit-variance to stabilize training.
We set $C'=64$ in \abbself and $C_l=16$ in \abbcross module.
For the $N$-way $K$-shot evaluation, we test 15 query samples for each class in an episode and report average classification accuracy with 95\% confidence intervals of randomly sampled 2,000 test episodes.
The hyperparameter $\lambda$ is set to 0.25, 0.5, 1.5 for ImageNet derivatives, CIFAR-FS, CUB, respectively.
$\gamma$ is set to 2 for CUB and 5 otherwise.
We set $\tau=0.2, U,V=5$ in our experiments.

\input{tables/sota_table.tex}

\subsection{Comparison to the state-of-the-art methods}
Tables~\ref{table:sota1} and \ref{table:sota2} compare \ours and current few-shot classification methods on four datasets.
Our model uses a smaller backbone than that of several methods~\cite{qiao2018few, wDAE, leo, s2m2} yet sets a new state of the art in both 5-way 1-shot and 5-shot settings on \itmini ImageNet, CUB-200-2011, and CIFAR-FS datasets while being comparable to DeepEMD~\cite{deepemd} on \ittiered ImageNet.
Note that DeepEMD iteratively performs back-propagation steps at each inference, which is very slow; it takes 8 hours to evaluate 2,000 5-way 5-shot episodes while ours takes 1.5 minutes on the same machine with an Intel i7-7820X CPU and an NVIDIA TitanXp GPU.
We also find \ours outperforms transfer learning methods~\cite{closer, simpleshot, negmargin, rfs} that are not explicitly designed to learn cross-relation between a query and supports.
However, \ours benefits from explicitly meta-learning cross-image relations and is able to better recognize image relevance adaptively to given few-shot examples.

\begin{figure}[t!]
\centering
\scalebox{0.90}{
\includegraphics[width=1\linewidth]{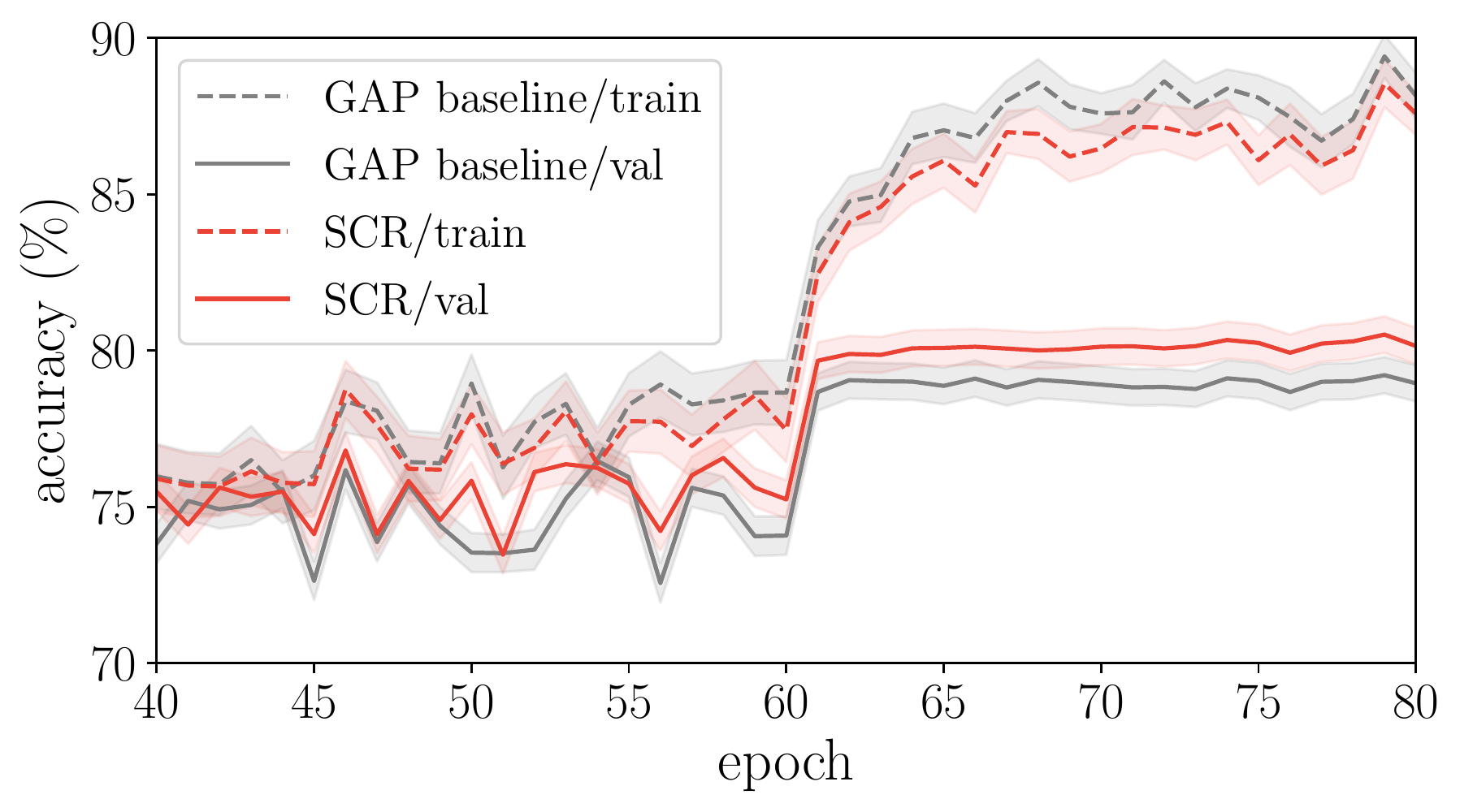}
}
\caption{
Learning curves of the GAP baseline and \abbself in terms of accuracy (\%) with 95 \% confidence intervals on CUB-200-2011.
The curves for the first 40 epochs are omitted.
\label{fig:overfitting}}
\end{figure}

\begin{figure*}[t!]
    \begin{minipage}[t]{.68\textwidth}
        \begin{minipage}[t]{.49\textwidth}
        \centering
        \scalebox{0.8}{
            \begin{tabular}[b]{cccc}
            \toprule
            \multirow{2}{*}{\abbself} & \multirow{2}{*}{\abbcross} & \small{\itmini -} & \multirow{2}{*}{CUB}  \\  
                                &                     & \small{ImageNet}    &   \\  
            \midrule
            \xmark & \xmark & 65.33 & 77.54 \\
            \cmark & \xmark & 66.66 \textcolor{red}{(+1.33)} & 78.69 \textcolor{red}{(+1.15)} \\
            \xmark & \cmark & 65.90 \textcolor{red}{(+0.57)} & 78.49 \textcolor{red}{(+0.95)} \\ \midrule
            \ccol \cmark & \ccol \cmark & \ccol \textbf{67.60} \textcolor{red}{(+2.27)} & \ccol \textbf{79.49}  \textcolor{red}{(+1.95)}\\
            \bottomrule 
            \end{tabular}
        }
        \captionof{table}{Effects of SCR and CCA. \label{table:ss_cs}}
        \end{minipage}
        \hfill    
        \begin{minipage}[t]{.5\textwidth}
            \centering
            \scalebox{0.8 }{
                \begin{tabular}[b]{lcccc}
                \toprule
                \multirow{2}{*}{ id }  &  \multirow{2}{*}{CCA}  & $h(\cdot)$ channel sizes & \small{\itmini -} & \multirow{2}{*}{CUB}  \\  
                                     &  & $C_\text{in} \rightarrow C_l \rightarrow C_\text{out}$  & \small{ImgNet}    &    \\  \midrule
                \texttt{(a)} & \xmark & \xmark & 65.33 & 77.54 \\
                \texttt{(b)} & \cmark & \xmark & 65.73 & 77.75 \\
                \texttt{(c)} & \cmark & 1 $\rightarrow$ \red{1} $\rightarrow$ 1 & 65.75 & 78.05 \\
                \texttt{(d)} & \cmark & \red{64} $\rightarrow$ 16 $\rightarrow$ 1 & 66.18 & 78.10 \\
                \midrule
                \ccol \texttt{(e)} & \ccol \cmark & \ccol 1 $\rightarrow$ 16 $\rightarrow$ 1 & \ccol \textbf{65.90} & \ccol \textbf{78.49} \\
                \bottomrule
                \end{tabular}
            }  
            \captionof{table}{Effects of \abbcross variants. \label{table:ca_design_choices}}
    \end{minipage} 
    \end{minipage} 
    \hfill
    \begin{minipage}[t]{.28\textwidth}
        \includegraphics[width=1\linewidth]{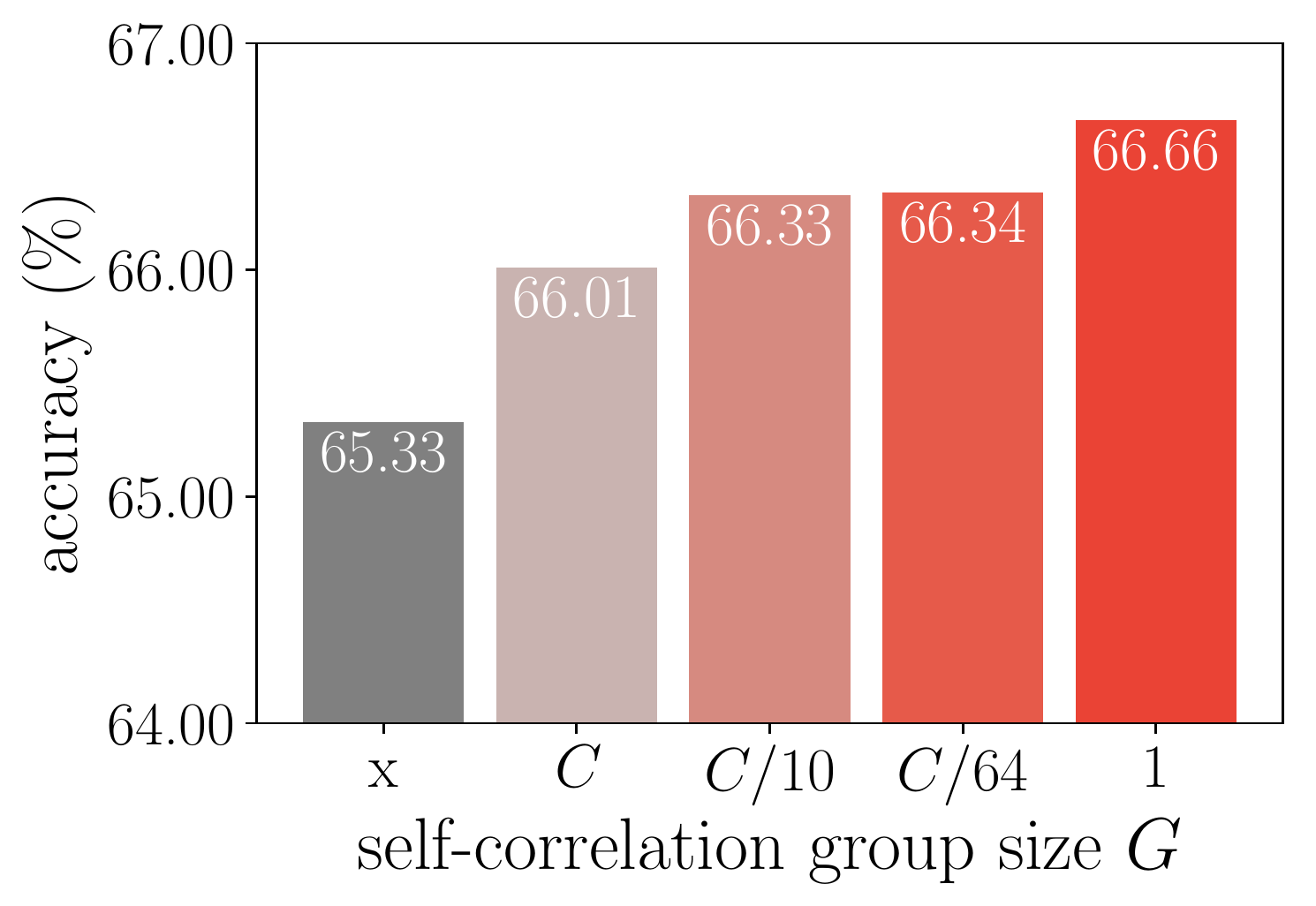}
        \captionof{figure}{Effects of the group size in SCR on \itmini ImageNet. \label{fig:ss_group}}
    \end{minipage} 
\vspace{-5mm}
\end{figure*}

\subsection{Ablation studies}
\label{sec:ablation}
To investigate the effects of core modules in \ours, we conduct extensive ablation studies either in the absence of each module or by replacing them with others and compare the results in the 5-way 1-shot setting.
For ease of comparison, we use a baseline model called GAP baseline that applies global-average pooling to base representations to obtain final embeddings.

\smallbreakparagraph{Effects of the proposed modules.}
Table~\ref{table:ss_cs} summarizes the effects of the \abbself and \abbcross modules.
Without \abbself, the model skips self-correlational learning, replacing its output $\bF$ with the base representation $\bZ$.
Without \abbcross, the model skips computing cross-correlation and obtains final image embeddings by simply averaging either $\bZ$ or $\bF$.
Both modules consistently improve classification accuracies on both datasets.
From the results, we observe that the effectiveness of \abbcross is more solid on CUB than that on \itmini ImageNet.
As the \abbcross module provides co-attention from the geometric consensus in cross-correlation patterns, it is particularly beneficial for a task where objects across different classes exhibit small geometric variations.
We also experimentally show that the self-correlational representation generalizes well to unseen classes than the base representation does as seen in Fig.~\ref{fig:overfitting}; the \abbself achieves lower training accuracy but higher validation accuracy than the GAP baseline.

\smallbreakparagraph{Design choices of \abbself.} 
To see the effectiveness of channel-wise correlation in \abbself, we replace the Hadamard product in Eq.~(\ref{eq:self_computation}) with group-wise cosine similarity in computing a self-correlation $\bR \in \Real^{H \times W \times U \times V \times C / G}$ and interpolate the group size $G$.
Namely, a group size $G>1$ compresses the channels of self-correlation, and $G=1$ becomes equivalent to the proposed method.
Figure~\ref{fig:ss_group} shows that the self-correlation with $G=C$, which represents the feature relation as a similarity scalar, is already effective, and further, the performance gradually increases as smaller group sizes are used; 
the model benefits from relational information, and the effect becomes greater with richer relation in the channel-wise correlation as similarly observed in \cite{zhao2020exploring}.

\smallbreakparagraph{Design choices of \abbcross.} 
We vary the components in the \abbcross module and denote the variants from \texttt{(b)} to \texttt{(d)} in Table~\ref{table:ca_design_choices} to verify our design choice.
In this study, we exclude \abbself learning to focus on the impact of the \abbcross.
We first examine a non-parametric baseline \texttt{(b)} by ablating all learnable parameters in the \abbcross module, \ie, we replace $\hat{\bC}$ in Eq.~(\ref{eq:attn}) with the cross-correlation between $\bZ_\textq$ and $\bZ_\texts$.
It shows marginal improvement from the GAP baseline \texttt{(a)}, which implies that the na\"ive cross-correlation hardly gives reliable co-attention maps.
Another variant \texttt{(c)} validates that the hidden channel dimension $C_l$ (Fig.~\ref{fig:cross_architecture}) helps the model capture diverse cross-correlation patterns.
The last variant \texttt{(d)} constructs cross-correlation preserving the channel dimension using Hadamard product instead of cosine similarity in Eq.~(\ref{eq:cross_correlation}).
Although it provides much information to the module and requires more learnable parameters (\texttt{(d)}: 797.3K {\em vs.}\  \texttt{(e)}: 45.8K), it is not very effective than the proposed one \texttt{(e)} possibly because too abundant correlations between two independent images negatively affect model generalization.

\begin{figure*}[t!]
\centering
\scalebox{1.0}{
\includegraphics[width=1\linewidth]{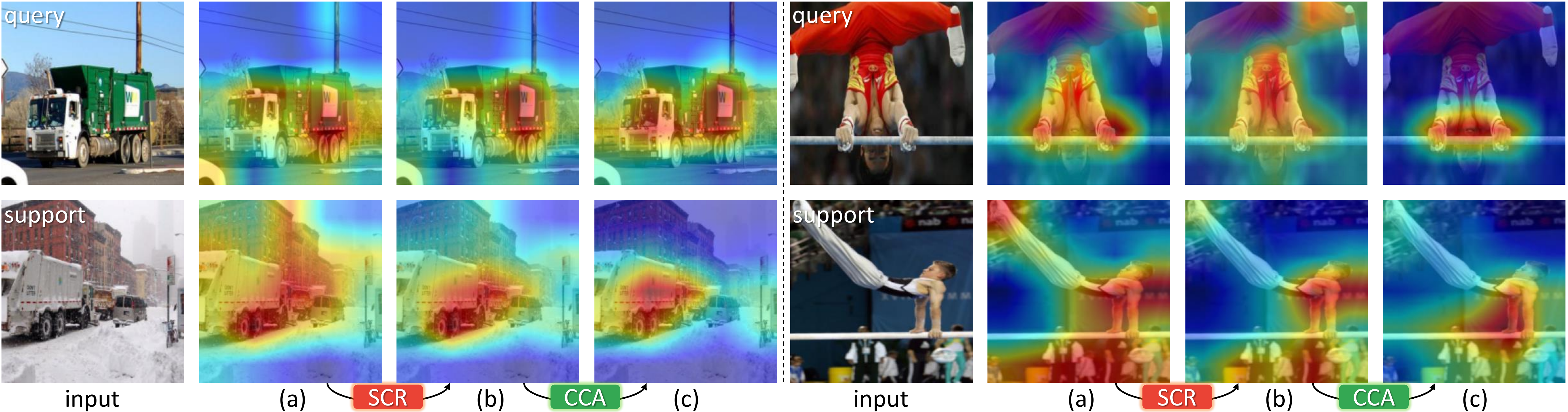}
}
\vspace{-5mm}
\caption{\textbf{Effects of \ours}.
\textbf{(a)}: Channel activation of base representation.
\textbf{(b)}: Channel activation of \abbself.
\textbf{(c)}: Attention map of \abbcross.
\label{fig:qual_main}}
\vspace{-3mm}
\end{figure*}

\begin{figure}[t!]
\centering
\scalebox{1.0}{
\includegraphics[width=1\linewidth]{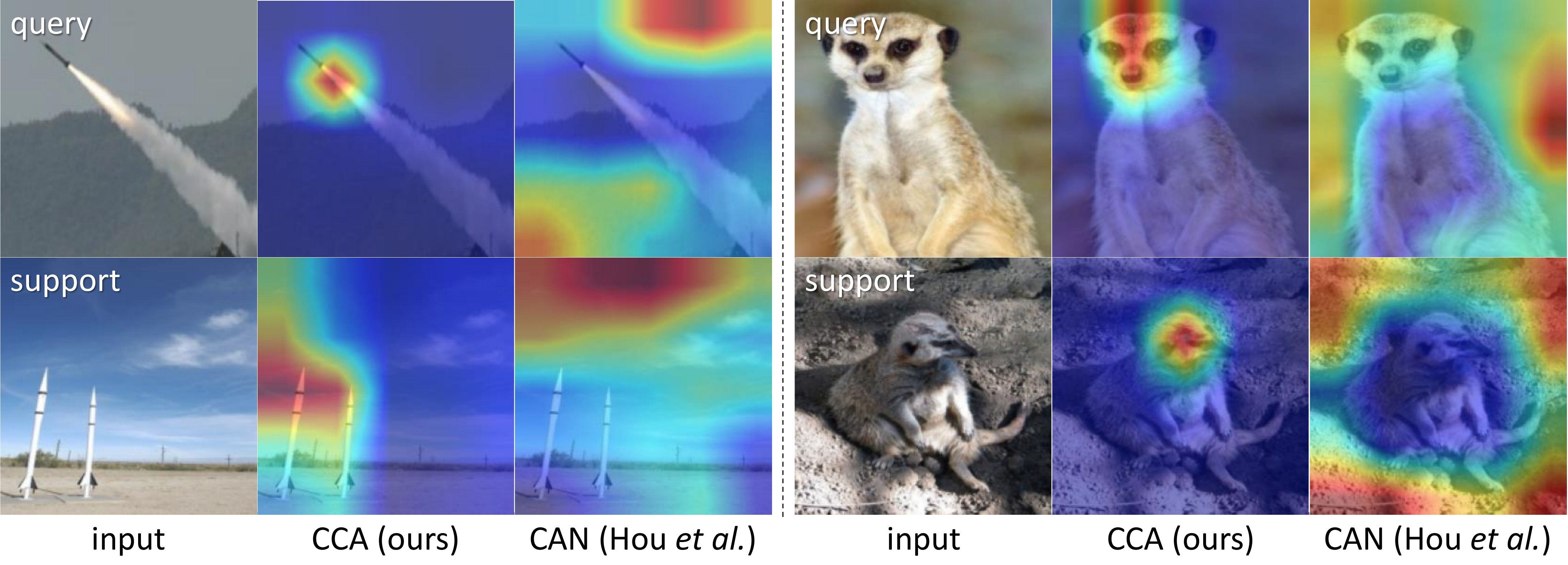}
}
\caption{Co-attention comparison with CAN~\cite{can}. 
Our CCA better attends to common objects against confusing backgrounds.
\label{fig:can}}
\vspace{-6mm}
\end{figure}

\input{tables/comparison_others.tex}

\subsection{Comparison with other attention modules}
In Table~\ref{table:comparison_other_methods}, we compare the proposed modules with other attention modules by replacing ours with others.
We first compare self-attention methods~\cite{nlsa, lsa, se} that attend to \textit{appearance} features based on feature similarity, while our \abbself module extracts \textit{relational} features based on local self-correlation.
In the comparison, \abbself outperforms self-attention methods, suggesting the effectiveness of learning self-correlation patterns for few-shot learning.
We find that learning such relational patterns of ``how each feature correlates with its neighbors'' is transferable to unseen classes and compensates the lack of data issue of few-shot learning.
While \abbself outperforms most methods, it closely competes with SCE~\cite{dccnet} on CUB.
SCE computes cosine similarity between a reference position and its neighbors and concatenates their similarities at the channel dimension in a fixed order.
SCE is powerful on the CUB dataset (77.54\% $\rightarrow$ 78.43\%) that has relatively little pose variation across images, however, it is disadvantageous on the \itmini ImageNet dataset (65.33\% $\rightarrow$ 63.39\%).
This is because SCE imprints the positional order in channel indices, which limits observing diverse neighborhood relations, whereas \abbself uses multiple channels to capture various relations with neighbors.

In Table~\ref{table:comparison_other_methods}, we also observe that \abbcross performs better than CAN~\cite{can} as well as other self-attention methods with a reasonable amount of additional parameters.
CAN first averages a 4D cross-correlation to a 2D correlation tensor and feeds it to multi-layer perceptrons that produce an attention mask, which is repeated similarly by switching the query and the support to generate co-attention maps.
We empirically find that the process of averaging the initial 4D correlation collapses fine and crucial match details between images.
Figure~\ref{fig:can} shows two examples that CAN is overwhelmed by dominant backgrounds and hardly attends to small objects.
Whereas, the \abbcross module updates cross-correlations while retaining the spatial dimensions hence successfully attends to relevant objects.

Combining the \abbself and the \abbcross modules, our model outperforms all the other methods.

\subsection{Qualitative results}
\label{sec:qual}
The relational embedding process and its attentional effects are shown in Figs.~\ref{fig:teaser} and~\ref{fig:qual_main}.
The columns (a) and (b) visualize the averaged channel activation of the base representation $\bZ$ and the self-correlational representation $\bF$, respectively.
The column (c) visualizes the 2D attention map $\bA$.
The images are randomly sampled from the \itmini ImageNet validation set, and activations are bi-linearly interpolated to the input image size.
The results demonstrate that the \abbself module can deactivate irrelevant features via learning self-correlation with neighborhood, \eg, the activation of a building behind a truck decreases.
The subsequent \abbcross module generates co-attention maps that focus on the common context between a query and a support, \eg, the hands grasping the bars are co-attended.

%% file: tables/sota_table.tex
\begin{table}[t!]
\centering
\tabcolsep 4pt
\small

\begin{subtable}{.5\textwidth}
\scalebox{0.96}{
\begin{tabular}{lccc}
\toprule
method & backbone & 5-way 1-shot & 5-way 5-shot\\  \midrule
\emph{cosine} classifier~\cite{closer} & \emph{ResNet12}  & 55.43 $\pm$ 0.81 & 77.18 $\pm$ 0.61 \\
TADAM~\cite{tadam}  & \emph{ResNet12}  & 58.50 $\pm$ 0.30 & 76.70 $\pm$ 0.30 \\
Shot-Free~\cite{shotfree}  & \emph{ResNet12}  & 59.04   & 77.64   \\
TPN~\cite{tpn}  & \emph{ResNet12}  & 59.46   & 75.65  \\
PPA~\cite{qiao2018few}  & \emph{WRN\small{-28-10}}${}^{\dag}$  & 59.60 $\pm$ 0.41  & 73.74  $\pm$ 0.19 \\
wDAE-GNN~\cite{wDAE}  & \emph{WRN\small{-28-10}}${}^{\dag}$  & 61.07 $\pm$ 0.15 & 76.75 $\pm$ 0.11 \\
MTL~\cite{mtl}  & \emph{ResNet12}  & 61.20 $\pm$ 1.80 & 75.50 $\pm$ 0.80 \\
LEO~\cite{leo}  & \emph{WRN\small{-28-10}}${}^{\dag}$  & 61.76 $\pm$ 0.08 & 77.59 $\pm$ 0.12\\
RFS-simple~\cite{rfs} & \emph{ResNet12}  & 62.02 $\pm$ 0.63 & 79.64 $\pm$ 0.44\\
DC~\cite{dc}  & \emph{ResNet18}  & 62.53 $\pm$ 0.19 & 79.77 $\pm$ 0.19\\
ProtoNet~\cite{protonet}  & \emph{ResNet12}  & 62.39 $\pm$ 0.21  & 80.53 $\pm$ 0.14 \\ %
MetaOptNet~\cite{metaoptnet}  & \emph{ResNet12}  & 62.64 $\pm$ 0.82 & 78.63 $\pm$ 0.46\\
SimpleShot~\cite{simpleshot}  & \emph{ResNet18}  & 62.85 $\pm$ 0.20 & 80.02 $\pm$ 0.14\\
MatchNet~\cite{matchingnet}  & \emph{ResNet12}  & 63.08 $\pm$ 0.80 & 75.99 $\pm$ 0.60\\
S2M2~\cite{s2m2} & \emph{ResNet34}${}^{\dag}$  & 63.74 $\pm$ 0.18 & 79.45  $\pm$ 0.12 \\
CAN~\cite{can}  & \emph{ResNet12}  & 63.85 $\pm$ 0.48 & 79.44 $\pm$ 0.34\\
NegMargin~\cite{negmargin}  & \emph{ResNet12}  & 63.85 $\pm$ 0.81 & 81.57 $\pm$ 0.56\\
CTM~\cite{ctm}  & \emph{ResNet18}  & 64.12 $\pm$ 0.82 & 80.51 $\pm$ 0.13\\
DeepEMD~\cite{deepemd}  & \emph{ResNet12}  & 65.91 $\pm$ 0.82 & 82.41 $\pm$ 0.56 \\  
FEAT~\cite{feat}  & \emph{ResNet12}  & 66.78 $\pm$ 0.20 & 82.05 $\pm$ 0.14\\\midrule
\ccol \textbf{\ours (ours)}  & \ccol \emph{ResNet12}  & \ccol \textbf{67.60 $\pm$ 0.44} & \ccol \textbf{82.58 $\pm$ 0.30} \\ \bottomrule 
\end{tabular}
}
\caption{\small Results on \emph{mini}ImageNet dataset.}
\end{subtable}
\begin{subtable}{.5\textwidth}
\scalebox{0.96}{
\begin{tabular}{lccc}
\toprule
method & backbone & 5-way 1-shot & 5-way 5-shot\\  \midrule
\emph{cosine} classifier~\cite{closer} & \emph{ResNet12}  & 61.49 $\pm$ 0.91  & 82.37 $\pm$ 0.67 \\
Shot-Free~\cite{shotfree}  & \emph{ResNet12}  &  63.52   &  82.59   \\
TPN~\cite{tpn}  & \emph{ResNet12}  & 59.91 $\pm$ 0.94  & 73.30 $\pm$ 0.75 \\
PPA~\cite{qiao2018few}  & \emph{WRN\small{-28-10}}${}^{\dag}$  & 65.65 $\pm$ 0.92  & 83.40 $\pm$ 0.65\\
wDAE-GNN~\cite{wDAE}  & \emph{WRN\small{-28-10}}${}^{\dag}$  & 68.18 $\pm$ 0.16  & 83.09 $\pm$ 0.12 \\
LEO~\cite{leo}  & \emph{WRN\small{-28-10}}${}^{\dag}$  & 66.33 $\pm$ 0.05  & 81.44 $\pm$ 0.09 \\
MetaOptNet~\cite{metaoptnet}  & \emph{ResNet12}  & 65.99 $\pm$ 0.72  & 81.56 $\pm$ 0.53 \\
ProtoNet~\cite{protonet}  & \emph{ResNet12}  & 68.23 $\pm$ 0.23  & 84.03 $\pm$ 0.16 \\ %
MatchNet~\cite{matchingnet}  & \emph{ResNet12}  & 68.50 $\pm$ 0.92 & 80.60 $\pm$ 0.71 \\
CTM~\cite{ctm}  & \emph{ResNet18}  & 68.41 $\pm$ 0.39  & 84.28 $\pm$ 1.73 \\
RFS-simple~\cite{rfs} & \emph{ResNet12}  & 69.74 $\pm$ 0.72 & 84.41 $\pm$ 0.55\\
CAN~\cite{can}  & \emph{ResNet12}  & 69.89 $\pm$ 0.51 & 84.23 $\pm$ 0.37\\
FEAT~\cite{feat}  & \emph{ResNet12}  & 70.80 $\pm$ 0.23 & 84.79 $\pm$ 0.16\\
DeepEMD~\cite{deepemd}  & \emph{ResNet12}  & 71.16 $\pm$ 0.87  & \textbf{86.03 $\pm$ 0.58} \\  \midrule
\ccol \textbf{\ours (ours)}  & \ccol \emph{ResNet12}  & \ccol \textbf{71.61 $\pm$ 0.51} & \ccol 85.28 $\pm$ 0.35 \\ \bottomrule
\end{tabular}
}
\caption{\small Results on \emph{tiered}ImageNet dataset.}
\end{subtable}
\caption{Comparison with the state-of-the-art 5-way 1-shot and 5-way 5-shot accuracy (\%) with 95\% confidence intervals on \textbf{(a)} \emph{mini}ImageNet and \textbf{(b)} \emph{tiered}ImageNet. 
``$\dag$'' denotes larger backbones than \emph{ResNet12}. }
\label{table:sota1}	
\vspace{-2mm}
\end{table}

\begin{table}[t!]
\centering
\tabcolsep 4pt
\small
\centering
\begin{subtable}{.5\textwidth}
\scalebox{0.98}{
\begin{tabular}{lccc}
\toprule
method & backbone & 5-way 1-shot & 5-way 5-shot\\  \midrule
ProtoNet~\cite{protonet} & \emph{ResNet12} & 66.09 $\pm$ 0.92 & 82.50 $\pm$ 0.58 \\
RelationNet~\cite{relationnet} & \emph{ResNet34}${}^{\dag}$ & 66.20 $\pm$ 0.99 & 82.30 $\pm$ 0.58 \\	%
MAML~\cite{relationnet} & \emph{ResNet34}${}^{\dag}$ & 67.28 $\pm$ 1.08 & 83.47 $\pm$ 0.59 \\	%
\emph{cosine} classifier~\cite{closer} & \emph{ResNet12}  & 67.30 $\pm$ 0.86  & 84.75 $\pm$ 0.60 \\ %
MatchNet~\cite{matchingnet} & \emph{ResNet12} & 71.87 $\pm$ 0.85 & 85.08 $\pm$ 0.57 \\
NegMargin~\cite{negmargin}  & \emph{ResNet18} & 72.66 $\pm$ 0.85 & 89.40 $\pm$ 0.43 \\
S2M2~\cite{s2m2} & \emph{ResNet34}${}^{\dag}$  & 72.92 $\pm$ 0.83 & 86.55 $\pm$ 0.51 \\
FEAT*~\cite{feat}  & \emph{ResNet12}  & 73.27 $\pm$ 0.22 & 85.77 $\pm$ 0.14\\
DeepEMD~\cite{deepemd}  & \emph{ResNet12} &  75.65 $\pm$ 0.83 &  88.69 $\pm$ 0.50 \\ \midrule
\ccol \textbf{\ours (ours)}  & \ccol \emph{ResNet12}  & \ccol \textbf{79.49 $\pm$ 0.44} & \ccol \textbf{91.11 $\pm$ 0.24} \\ \bottomrule
\end{tabular}
}
\caption{\small Results on CUB-200-2011 dataset.}
\end{subtable}

\centering
\begin{subtable}{.5\textwidth}
\scalebox{0.95}{
\begin{tabular}{lccc}
\toprule
method & backbone & 5-way 1-shot & 5-way 5-shot\\  \midrule
\emph{cosine} classifier~\cite{closer} & \emph{ResNet34}${}^{\dag}$  & 60.39 $\pm$ 0.28 & 72.85 $\pm$ 0.65 \\ %
S2M2~\cite{s2m2} & \emph{ResNet34}${}^{\dag}$  & 62.77 $\pm$ 0.23 & 75.75 $\pm$ 0.13 \\
Shot-Free~\cite{shotfree} & \emph{ResNet12}  & 69.2 & 84.7 \\
RFS-simple~\cite{rfs}  & \emph{ResNet12} & 71.5 $\pm$ 0.8  & 86.0 $\pm$ 0.5 \\
ProtoNet~\cite{protonet} & \emph{ResNet12}  & 72.2 $\pm$ 0.7  & 83.5 $\pm$ 0.5 \\ %
MetaOptNet~\cite{metaoptnet} & \emph{ResNet12}  & 72.6 $\pm$ 0.7  & 84.3 $\pm$ 0.5\\
Boosting~\cite{boosting} & \emph{WRN-28-10}${}^{\dag}$  & 73.6 $\pm$ 0.3  & 86.0 $\pm$ 0.2 \\ \midrule
\ccol \textbf{\ours (ours)}  & \ccol \emph{ResNet12}  & \ccol \textbf{74.51 $\pm$ 0.46} & \ccol \textbf{86.60 $\pm$ 0.32} \\ \bottomrule 
\end{tabular}
}
\caption{\small Results on CIFAR-FS dataset.}
\end{subtable}
\caption{Comparison with the state-of-the-art 5-way 1-shot and 5-way 5-shot accuracy (\%) with 95\% confidence intervals on \textbf{(a)} CUB-200-2011 and \textbf{(b)} CIFAR-FS.
``$\dag$'' denotes larger backbones than \emph{ResNet12}, and ``*'' denotes reproduced one.}

\label{table:sota2}	
\vspace{-2mm}
\end{table}

%% file: tables/comparison_others.tex
\begin{table}[t!]
    \centering
    \scalebox{0.82}{
    \begin{tabular}[b]{lccccr}
    \toprule
    \multirow{2}{*}{method} & \multirow{2}{*}{self} & \multirow{2}{*}{cross} & \small{\itmini -}   & \multirow{2}{*}{CUB}  & \# add.  \\  
                            &                       &                        & \small{ImgNet}  &                       &  params.  \\  
    \midrule
    \textbf{GAP baseline} & \xmark & \xmark                         & 65.38 & 77.54 & 0K \\
    \textbf{SE~\cite{se}} & \cmark (self-attn) & \xmark             & 63.34 & 78.40 & 83.8K \\
    \textbf{non-local~\cite{nlsa}} & \cmark (self-attn) & \xmark    & 65.00 & 77.11 & 822.1K \\
    \textbf{local~\cite{lsa}} & \cmark (self-attn) & \xmark         & \underline{66.26} & 78.19 & 1644.1K \\
    \textbf{SCE~\cite{dccnet}} & \cmark (self-sim) & \xmark         & 63.39 & \underline{78.43} & 89.2K \\ 
    \textbf{CAN*~\cite{can}} & \xmark & \cmark                      & 65.66 & 77.77 & 0.3K \\ 
    \midrule
    \ccol \textbf{\abbself}  & \ccol \cmark (self-corr) & \ccol \xmark & \ccol 66.66 & \ccol 78.69 & \ccol 157.3K \\
    \ccol \textbf{\abbcross} & \ccol \xmark & \ccol \cmark  & \ccol 66.00 & \ccol 78.49 & \ccol 45.8K \\
    \ccol \textbf{\abbself + \abbcross} & \ccol \cmark (self-corr) & \ccol \cmark & \ccol \textbf{67.60} & \ccol \textbf{79.49} & \ccol 203.2K \\
    \bottomrule
    \end{tabular}
    }
    \caption{
    Accuracy (\%) and the number of additional learnable parameters of other relation-based methods. 
    ``*'' denotes reproduced one under a controlled environment for a fair comparison, and underline denotes the best performance among others.
    \label{table:comparison_other_methods}}
\vspace{-5mm}
\end{table}

%% file: sections/6_conclusion.tex
\section{Conclusion}
In this work, we have proposed the relational embedding network for few-shot classification, which leverages the self-correlational representation and the cross-correlational attention. 
Combining the two modules, our method has achieved the state of the art on the four standard benchmarks.
One of our experimental observations is that self-attention mechanism~\cite{nlsa, lsa} is prone to overfitting to the training set so that it does not generalize to unseen classes in the few-shot learning context.
Our work, however, has shown that learning structural correlations between visual features better generalizes to unseen object classes and brings performance improvement to few-shot image recognition, suggesting a promising direction of relational knowledge as a transferable prior.

%% file: sections/7_acknowledgements.tex
\paragraph{Acknowledgements.}
This work was supported by Samsung Electronics Co., Ltd. (IO201208-07822-01) and the IITP grants (No.2019-0-01906, AI Graduate School Program - POSTECH) (No.2021-0-00537, Visual common sense through self-supervised learning for restoration of invisible parts in images) funded by Ministry of Science and ICT, Korea.

%% file: sections/8_supp.tex
\section{Appendix}
In this appendix, we provide additional details and results of our method.

\subsection{Alternative derivation of relational embedding}
Equations (\ref{eq:attn}), (\ref{eq:attentive_pooling_q}), and (\ref{eq:attentive_pooling_s}) in the main paper describe the process of deriving relational embeddings, $\bq$ and $\bs \in \Real^{C}$, using pre-computed co-attention maps, $\bA_\textq$ and $\bA_\texts \in \Real^{H \times W}$, where the attention maps themselves provide interpretable visualization, \eg, Fig.~\ref{fig:teaser}(c) in the main paper.
In this section, we derive $\bq$ and $\bs$ in an alternative way of not explicitly introducing the attention maps, $\bA_\textq$ and $\bA_\texts$, but \textit{multiplying a feature map by cross-correlation}, which is used in spatial attention work~\cite{sun2020mining, dualattentionnet, superglue}.
Let us denote the normalized cross-correlation tensor in Eq.~(\ref{eq:attn}) by 
\begin{eqnarray}
\tilde{\bC} = \frac{\exp{(\hat{\bC}(\bx_\textq, \bx_\texts) / \gamma)} }{\sum_{\bx_\textq'}
	\exp{(\hat{\bC}(\bx_\textq', \bx_\texts) / \gamma)}}
\end{eqnarray}
and reshape it to a 2D matrix: $\tilde{\bC} \in \Real^{HW \times HW}$.

The relational embedding $\bq$ is equivalently derived by multiplying two matrices $\tilde{\bC}^{\top}$ and $\bF_\textq \in \Real^{HW \times C}$ followed by average pooling:
\begin{align}
\nonumber
\bq &= \sum_{\bx_\textq} \left( \underbrace{\left( \frac{1}{HW} \sum_{\bx_\texts} \tilde{\bC}(\bx_\textq, \bx_\texts) \right)}_{\text{Eq.~(\ref{eq:attn})}} \bF_\textq(\bx_\textq) \right) \, (\text{Eq.~(\ref{eq:attentive_pooling_q})}) \\
\nonumber
    &= \frac{1}{HW} \sum_{\bx_\textq} \left( \sum_{\bx_\texts} \tilde{\bC}(\bx_\textq, \bx_\texts) \right) \bF_\textq(\bx_\textq) \\
\nonumber
    &= \frac{1}{HW} \sum_{\bx_\texts} \sum_{\bx_\textq}  \tilde{\bC}(\bx_\textq, \bx_\texts) \bF_\textq(\bx_\textq) \\
\nonumber
    &= \frac{1}{HW} \sum_{\bx_\texts} \underbrace{\sum_{\bx_\textq}  \tilde{\bC}^{\top}(\bx_\texts, \bx_\textq) \bF_\textq(\bx_\textq)}_{\text{matrix multiplication}} \\
    &= \frac{1}{HW} \sum_{\bx_\texts} \underbrace{\tilde{\bC}^{\top}\bF_\textq}_{\Real^{HW \times C}} (\bx_\texts).
\end{align}
Here, $\tilde{\bC}^{\top}\bF_\textq$ is considered as softly-aligning the query feature map $\bF_\textq$ in the light of each position of the support using the cross-correlation $\tilde{\bC}^{\top}$.

Likewise, the relational embedding $\bs$ is computed as 
\begin{eqnarray}
\bs = \frac{1}{HW} \sum_{\bx_\textq} \tilde{\bC}\bF_\texts(\bx_\textq).
\end{eqnarray}

\subsection{Comprehensive details on implementation}
For training, we use an SGD optimizer with a momentum of 0.9 and a learning rate of 0.1.
We train 1-shot models for 80 epochs and decay the learning rate by a factor of 0.05 at each \{60, 70\} epoch.
To train 5-shot models, we run 60 epochs and decay the learning rate at each \{40, 50\} epoch.
We randomly construct a training batch of size 128 for the ImageNet family~\cite{matchingnet, tieredimagenet} and 64 for CUB~\cite{cub} \& CIFAR-FS~\cite{cifarfs} to compute $\mathcal{L}_{\text{anchor}}$.
This objective is jointly optimized from scratch with $\mathcal{L}_{\text{metric}}$ and $\mathcal{L}_{\text{anchor}}$ as described in Sec.~\ref{sec:learning_objective}.
For a fair comparison, we adopt the same image sizes, the backbone network, the data augmentation techniques, and the embedding normalization following the recent work of \cite{deepemd, feat}.

\subsection{Ablation studies}
We provide more ablation studies on CUB~\cite{cub} and \itmini ImageNet~\cite{matchingnet} in the 5-way 1-shot setting.

\begin{table}[t!]
\centering
\scalebox{0.92}{
\begin{tabular}{lccc}
\toprule
self-correlation & category of & \itmini   & \multirow{2}{*}{CUB} \\
computation      & neighbors  & ImgNet    &  \\  \midrule
\xmark \, (GAP baseline) & \xmark & 65.33 & 77.54 \\
$\bR \in \Real^{H \times W \times \red{H} \times \red{W} \times C}$ & absolute & 66.41 & 76.34 \\  \midrule
\ccol $\bR \in \Real^{H \times W \times U \times V \times C}$  & \ccol relative & \ccol \textbf{66.66} & \ccol  \textbf{78.69} \\   \bottomrule
\end{tabular}
}
\caption{Comparison between absolute and relative neighborhood space in computing the self-correlation tensor $\bR$.}
\label{table:ablation_self_corr_computation}
\end{table}

\subsubsection{Self-correlation computation with relative {\em vs.}\ absolute neighbors}
We validate the importance of \textit{relative} neighborhood correlations of a self-correlation tensor $\bR$ in Table~\ref{table:ablation_self_corr_computation}.
We set $H \! = \! W \! = \! U \! = \! V$ such that the two models have the same input sizes for a fair comparison.
The results show the superiority of the relative neighborhood correlation.
An advantage of the relative correlation over the absolute one is that relative correlations provide a translation-invariant neighborhood space.
For example, let us consider a self-correlation between a reference position $\bx$ and its neighbors.
While an absolute correlation $(\bx, \bx') \in \Real^{H \times W \times H \times W}$ provides a variable neighborhood space as $\bx$ translates by $\bt$: $(\bx + \bt, \underline{\bx' + \bt})$, a relative correlation $(\bx, \bp) \in \Real^{H \times W \times U \times V}$ provides a consistent view of the neighborhood space no matter how $\bx$ moves: $(\bx + \bt, \underline{\bp})$.

\begin{table}[t!]
\centering
\scalebox{0.92}{
\begin{tabular}{lccc}
\toprule
4D convolution & \itmini & \multirow{2}{*}{CUB} & GPU time \\
kernels &  ImgNet &  & (\textit{ms}) \\ \midrule
\xmark \, (GAP baseline) & 65.33 & 77.54 & 27.74 \\
vanilla 4D~\cite{ncnet}  & 65.59 & 78.89 & 60.35 \\  \midrule
\ccol separable 4D~\cite{vcn} & \ccol \textbf{65.90} & \ccol \textbf{78.49} & \ccol 34.97  \\   \bottomrule
\end{tabular}
}
\caption{Comparison between 4D convolutions for $h(\cdot)$.
}
\label{table:4d}
\vspace{-2mm}
\end{table}

\subsubsection{Separable {\em vs.}\ vanilla 4D convolution on \abbcross}
Comparison between the original vanilla 4D convolutional kernels~\cite{ncnet} and separable 4D kernels~\cite{vcn} is summarized in Table~\ref{table:4d}, where we adopt the separable one for its efficiency.
Note that the separable 4D kernels approximate the vanilla $3 \! \times \! 3 \! \times \! 3 \! \times \! 3$ kernels by two sequential $3 \! \times \! 3 \! \times \! 1 \! \times \! 1$ and $1 \! \times \! 1 \! \times \! 3 \! \times \! 3$ kernels followed by a point-wise convolution.
The reported GPU time in Table~\ref{table:4d} is an average time for processing an episode and is measured using a CUDA event wrapper in PyTorch~\cite{pytorch}.
While the two kinds of kernels closely compete with each other in terms of accuracy, the separable one consumes less computational costs.

\begin{table}[t!]
\centering
\scalebox{0.92}{
\begin{tabular}{lcr}
\toprule
\multirow{2}{*}{method} & 5-way 1-shot & \# add. \\
& accuracy (\%) & params \\ \midrule
CAN~\cite{can}  & 63.85 $\pm$ 0.48 & 0.3K \\
\ccol \ours (ours)    & \ccol \textbf{67.60 $\pm$ 0.44} & \ccol 203.2K \\
LEO~\cite{leo}  & 61.76 $\pm$ 0.08 & 248.8K \\
CTM~\cite{ctm}  & 64.12 $\pm$ 0.82 & 305.8K \\
FEAT~\cite{feat}& 66.78 $\pm$ 0.20 & 1640.3K \\
MTL~\cite{mtl}  & 61.20 $\pm$ 1.80 & 4301.1K \\
wDAE~\cite{wDAE}& 61.07 $\pm$ 0.15 & 11273.2K \\ \bottomrule
\end{tabular}
}
\caption{Performance comparison in terms of model size and accuracy (\%) on \itmini ImageNet.}
\label{table:num_params}
\vspace{-4mm}
\end{table}

\subsubsection{Number of parameters}
We measure the number of additional model parameters of recent methods and compare them with \ours in Table~\ref{table:num_params}.
Table~\ref{table:num_params} studies the effect of {\em additional} parameters only so we collect publicly available codes of methods that use additional parameterized modules~\cite{can, leo, ctm, feat, mtl, wDAE}, and intentionally omit \cite{closer, metaoptnet, negmargin, s2m2, protonet, rfs, simpleshot, deepemd} as their trainable parameters are either in the backbone network or in the last fully-connected layer.
Compared to the largest model~\cite{wDAE}, ours performs significantly better (67.60 {\em vs.}~61.07) while introducing 55 times less additional capacity (203.2K {\em vs.}~11.2M).

\begin{figure}[t!]
\centering
\includegraphics[width=0.95\linewidth]{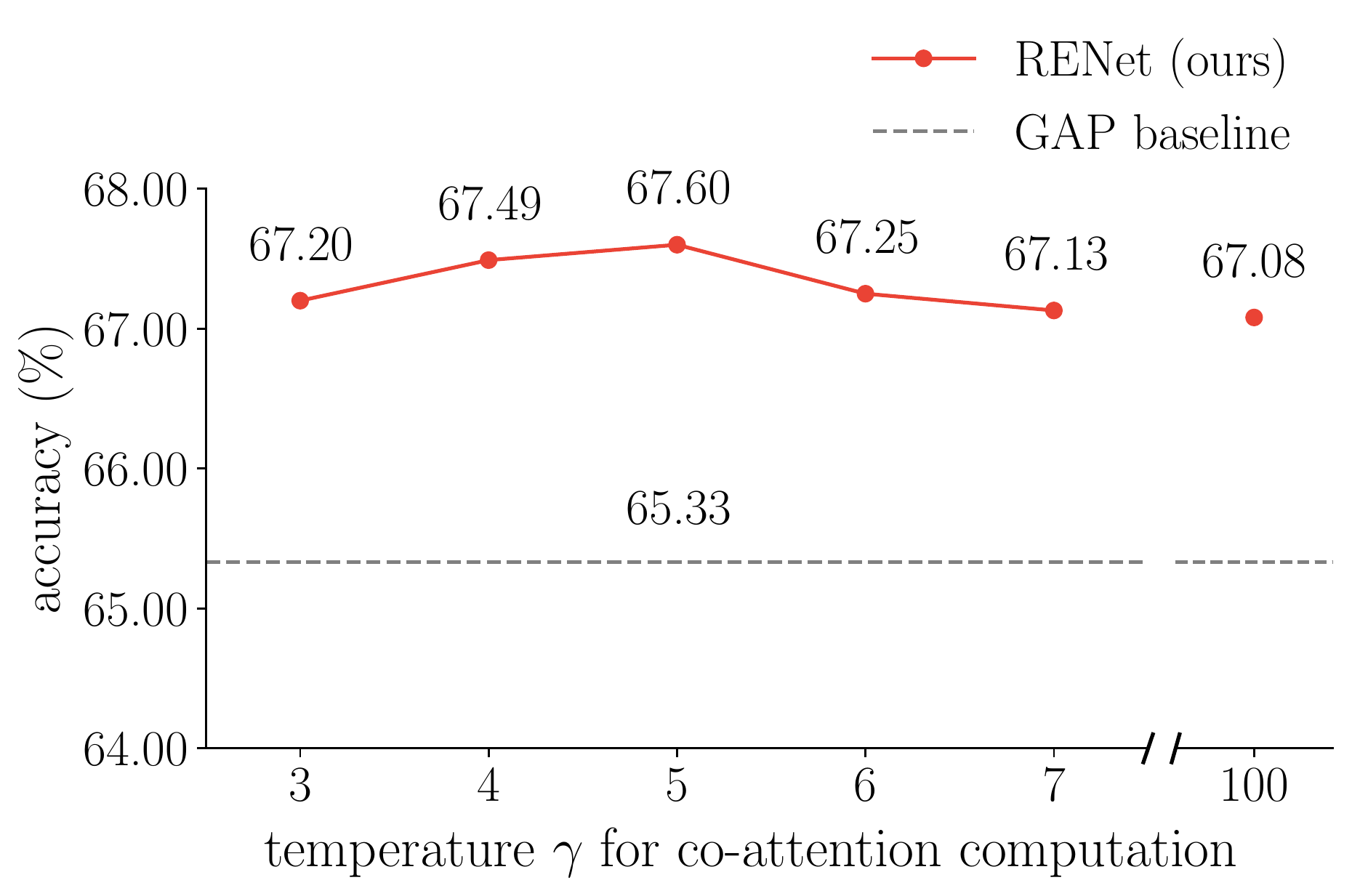}
\vspace{-2mm}
\caption{Accuracy (\%) of varying $\gamma$ on \itmini ImageNet.
\label{fig:temperature}}
\vspace{-2mm}
\end{figure}

\subsubsection{Temperature $\gamma$ for co-attention computation}
We investigate the impact of the hyper-parameter $\gamma$ that controls the smoothness of the output attention map (Eq.~(\ref{eq:attn})).
As its name ``temperature'' suggests, a higher temperature outputs a smoother attention map, while a lower temperature outputs a peakier one.
Figure~\ref{fig:temperature} shows that the temperature $\gamma$ has a certain point that maximizes the accuracy by appropriately balancing the smoothness factor.
Interestingly, an extremely high temperature $\gamma = 100$ degrades accuracy by making all attention scores evenly distributed.
It is noteworthy that our full model \ours with a range of $\gamma \in \{3, 4, 5, 6, 7\}$ outperforms all existing methods on the dataset.

\begin{figure}[t!]
\centering
\includegraphics[width=0.95\linewidth]{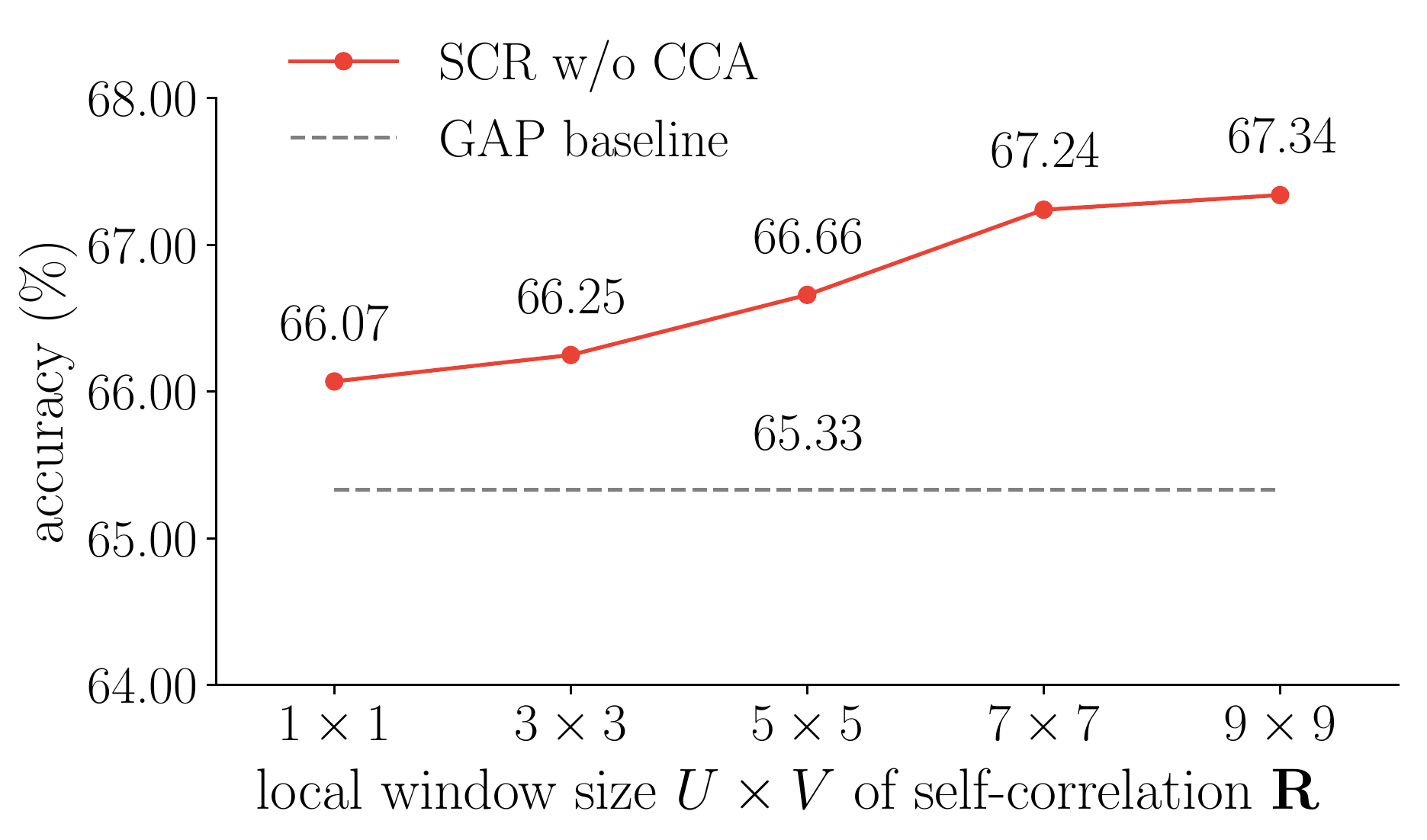}
\vspace{-2mm}
\caption{Accuracy (\%) of varying $U \times V$ on \itmini ImageNet.
\label{fig:uv}}
\vspace{-2mm}
\end{figure}

\begin{figure*}[t!]
\centering
\includegraphics[width=0.68\linewidth]{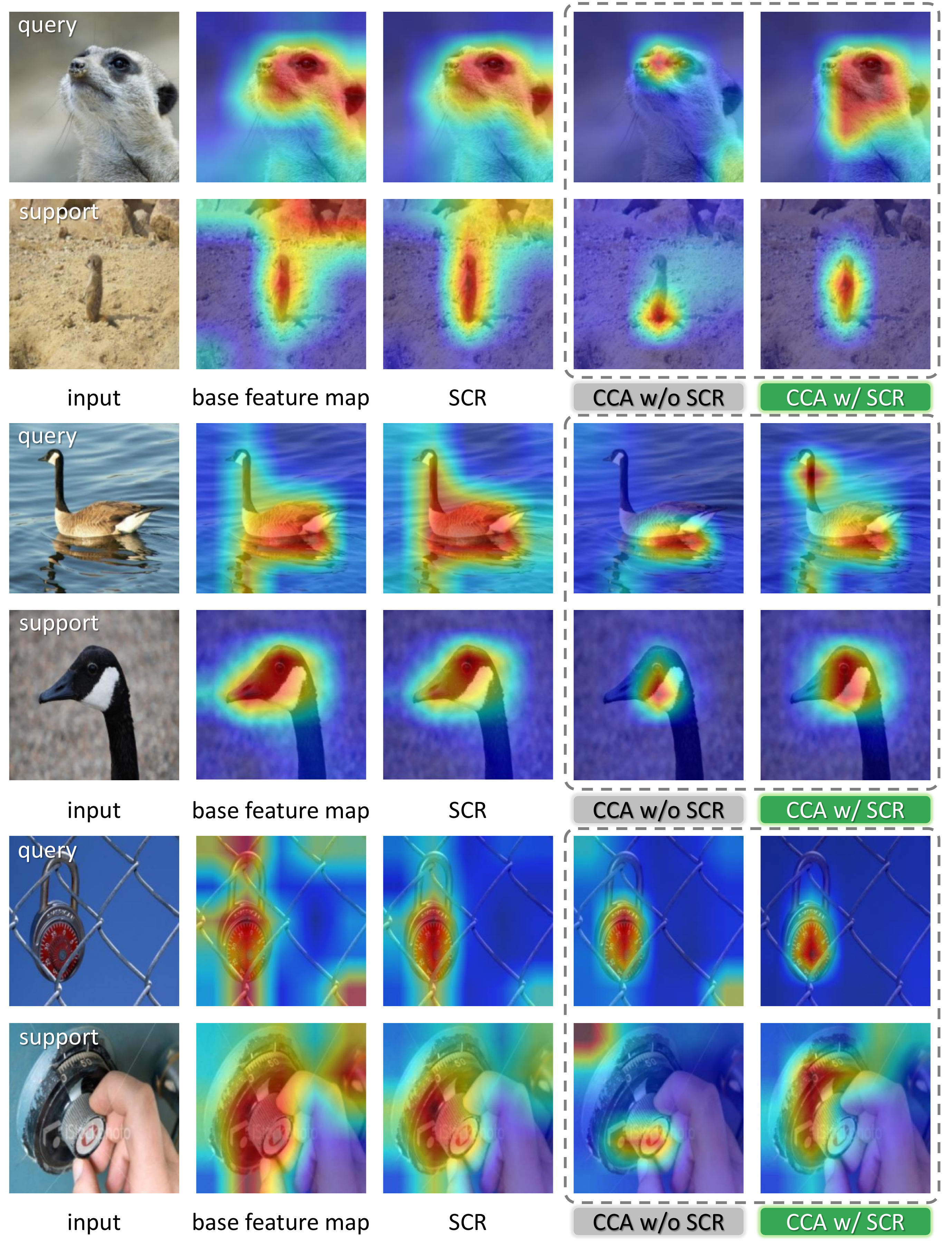}
\caption{\textbf{Effects of \abbself on \itmini ImageNet}.
``\abbcross w/ \abbself'' captures fine details between two images while ``\abbcross w/o \abbself'' often fails.
The ``base feature map'' and ``\abbself'' columns visualize average channel activations.
The ``\abbcross w/ \abbself'' and ``\abbcross w/o \abbself'' columns visualize co-attention maps.
\label{fig:effects_of_scr}}
\vspace{-2mm}
\end{figure*}

\subsubsection{Local window size $UV$ for \abbself}
To evaluate the effectiveness of learning relational features from local neighborhood correlation, we vary the local window size $UV$ of a self-correlation tensor $\bR \in \Real^{H \times W \times U \times V \times C}$.
As shown in Fig~\ref{fig:uv}, the accuracy steadily increases as more neighborhood correlations are learned, which indicates that learning relational structures is favorable for few-shot recognition.
Note that \abbself with $U = V = 1$ already outperforms the GAP baseline, which is an effect of learning from $l$2-normalized features (Eq.~\ref{eq:self_computation}).
Despite the consistent accuracy gain from observing wide local window, we choose $U = V = 5$ for all experiments \textit{to limit the space complexity} increased by a factor of $UV$.

\begin{figure*}[t!]
\centering
\includegraphics[width=0.9\linewidth]{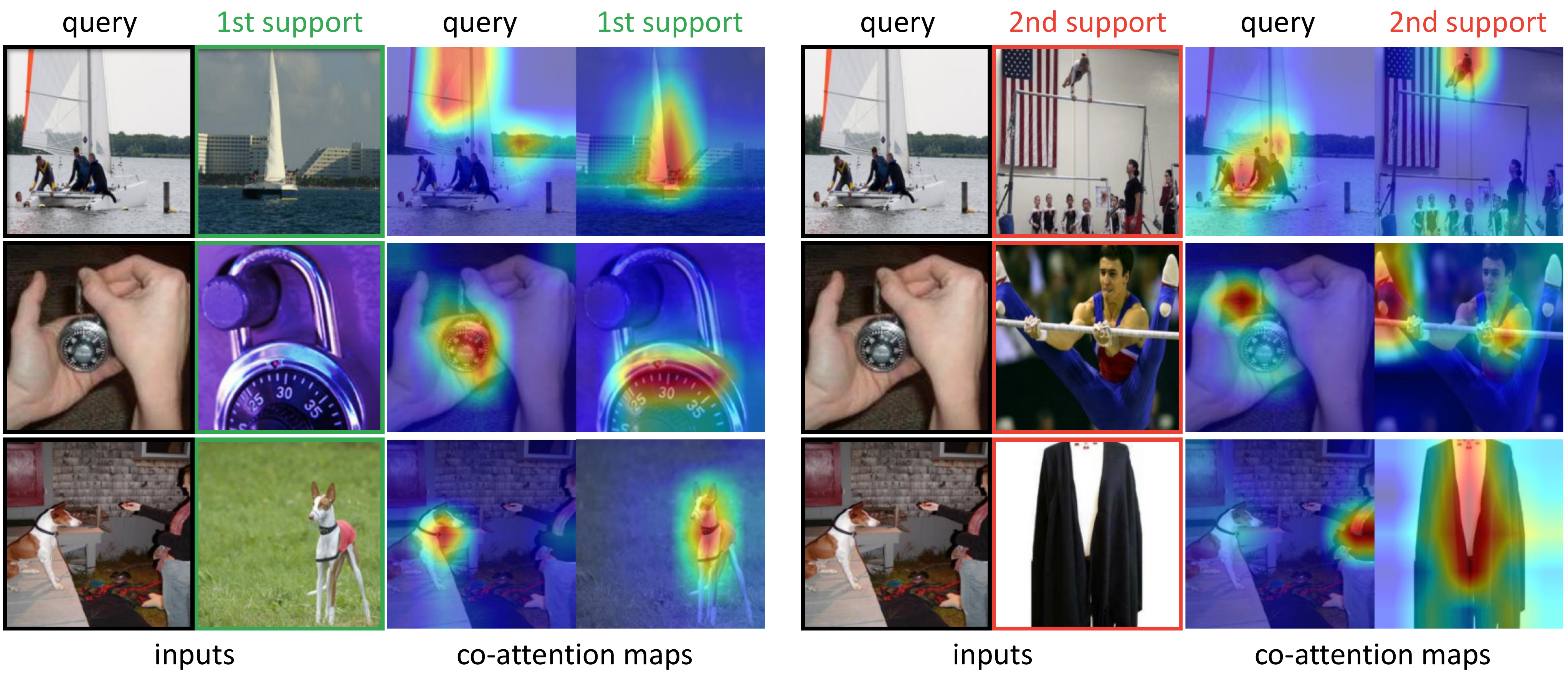}
\caption{\textbf{Co-attention maps on multi-object queries on \itmini ImageNet}.
The proposed \abbcross module can adaptively capture multiple objects in a query depending on the context of each support instance. 
\label{fig:multi_object_queries}}
\end{figure*}

\begin{figure*}[t!]
\centering
\includegraphics[width=0.783\linewidth]{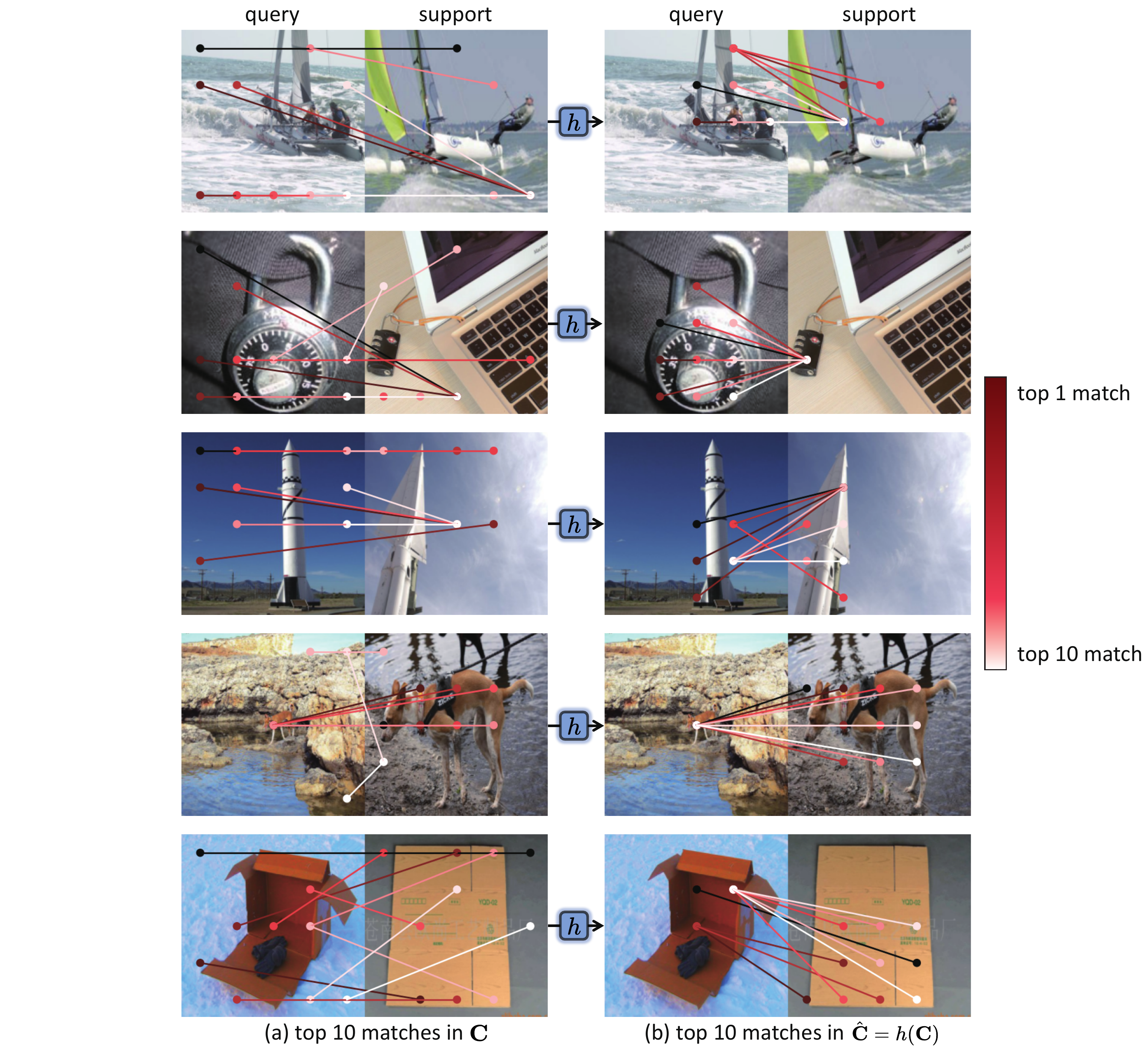}
\caption{\textbf{Visualization of cross-correlation on \itmini ImageNet}.
\textbf{(a)}: Top 10 matches in $\bC$ (initial cross-correlation). \textbf{(b)}: Top 10 matches in $\hat{\bC} = h(\bC)$ (refined cross-correlation).
Unreliable matches are filtered through $h(\cdot)$.
\label{fig:corr}}
\end{figure*}

\subsection{Qualitative results}
To demonstrate the effects of our method, we present additional qualitative results.
All images are sampled from the \itmini ImageNet validation set in the 5-way 1-shot setting.

\subsubsection{Effects of \abbself}
We ablate the \abbself module and demonstrate the effects of \abbself in Fig.~\ref{fig:effects_of_scr}.
The results show that ``CCA w/ SCR'' successfully attends to fine characteristics than ``CCA w/o SCR'' does, implying that the \abbself module provides reliable representation for the subsequent \abbcross module.

\subsubsection{Co-attention maps on multi-object queries}
Given a multi-object image as a query, we examine if the object regions can be adaptively highlighted depending on the support semantics in Fig.~\ref{fig:multi_object_queries}.
The \abbcross module successfully captures query regions that are semantically related with each support image.
This effect accords with the motivation of the \abbcross module, which is to adaptively provide ``where to attend'' between two image contexts.

\subsubsection{Cross-correlation refinement via $h(\cdot)$}
We demonstrate the effect of 4D convolutional block $h(\cdot)$ that filters out unreliable matches in the initial cross-correlation by analyzing neighborhood consensus patterns.
We visualize the top 10 matches among $2HW$ matching candidates computed by $\mathrm{argmax}$ of matching scores from each side.
As shown in Fig.~\ref{fig:corr}, the initial cross-correlation $\bC$ exhibits many spurious matches misled by indistinguishable appearance, \eg, matching two regions of the sky, whereas the updated cross-correlation $\hat{\bC}$ shows reliable and meaningful matches, \eg, matching two sails.

%% file: main.bbl
\begin{thebibliography}{10}\itemsep=-1pt

\bibitem{allen2019infinite}
Kelsey Allen, Evan Shelhamer, Hanul Shin, and Joshua Tenenbaum.
\newblock Infinite mixture prototypes for few-shot learning.
\newblock In {\em Proc. International Conference on Machine Learning (ICML)},
  2019.

\bibitem{bengio1990learning}
Yoshua Bengio, Samy Bengio, and Jocelyn Cloutier.
\newblock {\em Learning a synaptic learning rule}.
\newblock Citeseer, 1990.

\bibitem{cifarfs}
Luca Bertinetto, Joao~F Henriques, Philip Torr, and Andrea Vedaldi.
\newblock Meta-learning with differentiable closed-form solvers.
\newblock In {\em Proc. International Conference on Learning Representations
  (ICLR)}, 2018.

\bibitem{brendel2019approximating}
Wieland Brendel and Matthias Bethge.
\newblock Approximating cnns with bag-of-local-features models works
  surprisingly well on imagenet.
\newblock In {\em Proc. International Conference on Learning Representations
  (ICLR)}, 2019.

\bibitem{conceptlearners}
Kaidi Cao, Maria Brbic, and Jure Leskovec.
\newblock Concept learners for few-shot learning.
\newblock In {\em Proc. International Conference on Learning Representations
  (ICLR)}, 2021.

\bibitem{closer}
Wei-Yu Chen, Yen-Cheng Liu, Zsolt Kira, Yu-Chiang Wang, and Jia-Bin Huang.
\newblock A closer look at few-shot classification.
\newblock In {\em International Conference on Learning Representations (ICLR)},
  2019.

\bibitem{deselaers2010global}
Thomas Deselaers and Vittorio Ferrari.
\newblock Global and efficient self-similarity for object classification and
  detection.
\newblock In {\em 2010 IEEE Computer Society Conference on Computer Vision and
  Pattern Recognition}, 2010.

\bibitem{dhillon2019baseline}
Guneet~Singh Dhillon, Pratik Chaudhari, Avinash Ravichandran, and Stefano
  Soatto.
\newblock A baseline for few-shot image classification.
\newblock In {\em International Conference on Learning Representations}, 2019.

\bibitem{crosstransformers}
Carl Doersch, Ankush Gupta, and Andrew Zisserman.
\newblock Crosstransformers: spatially-aware few-shot transfer.
\newblock In {\em Advances in Neural Information Processing Systems (NeurIPS)},
  2020.

\bibitem{dosovitskiy2015flownet}
Alexey Dosovitskiy, Philipp Fischer, Eddy Ilg, Philip Hausser, Caner Hazirbas,
  Vladimir Golkov, Patrick Van Der~Smagt, Daniel Cremers, and Thomas Brox.
\newblock Flownet: Learning optical flow with convolutional networks.
\newblock In {\em Proc. IEEE International Conference on Computer Vision
  (ICCV)}, 2015.

\bibitem{fei2006one}
Li Fei-Fei, Rob Fergus, and Pietro Perona.
\newblock One-shot learning of object categories.
\newblock {\em IEEE Transactions on Pattern Analysis and Machine Intelligence
  (TPAMI)}, 28(4):594--611, 2006.

\bibitem{maml}
Chelsea Finn, Pieter Abbeel, and Sergey Levine.
\newblock Model-agnostic meta-learning for fast adaptation of deep networks.
\newblock In {\em Proc. International Conference on Machine Learning (ICML)},
  2017.

\bibitem{dualattentionnet}
Jun Fu, Jing Liu, Haijie Tian, Yong Li, Yongjun Bao, Zhiwei Fang, and Hanqing
  Lu.
\newblock Dual attention network for scene segmentation.
\newblock In {\em Proc. IEEE Conference on Computer Vision and Pattern
  Recognition (CVPR)}, 2019.

\bibitem{geirhos2018imagenet}
Robert Geirhos, Patricia Rubisch, Claudio Michaelis, Matthias Bethge, Felix~A
  Wichmann, and Wieland Brendel.
\newblock Imagenet-trained cnns are biased towards texture; increasing shape
  bias improves accuracy and robustness.
\newblock In {\em Proc. International Conference on Learning Representations
  (ICLR)}, 2019.

\bibitem{boosting}
Spyros Gidaris, Andrei Bursuc, Nikos Komodakis, Patrick P{\'e}rez, and Matthieu
  Cord.
\newblock Boosting few-shot visual learning with self-supervision.
\newblock In {\em Proc. IEEE International Conference on Computer Vision
  (ICCV)}, 2019.

\bibitem{gidaris2018dynamic}
Spyros Gidaris and Nikos Komodakis.
\newblock Dynamic few-shot visual learning without forgetting.
\newblock In {\em Proc. IEEE Conference on Computer Vision and Pattern
  Recognition (CVPR)}, 2018.

\bibitem{wDAE}
Spyros Gidaris and Nikos Komodakis.
\newblock Generating classification weights with gnn denoising autoencoders for
  few-shot learning.
\newblock In {\em Proc. IEEE Conference on Computer Vision and Pattern
  Recognition (CVPR)}, 2019.

\bibitem{resnet}
Kaiming He, Xiangyu Zhang, Shaoqing Ren, and Jian Sun.
\newblock Deep residual learning for image recognition.
\newblock In {\em Proc. IEEE Conference on Computer Vision and Pattern
  Recognition (CVPR)}, 2016.

\bibitem{hochreiter2001learning}
Sepp Hochreiter, A~Steven Younger, and Peter~R Conwell.
\newblock Learning to learn using gradient descent.
\newblock In {\em Proc. International Conference on Artificial Neural Networks
  (ICANN)}, 2001.

\bibitem{can}
Ruibing Hou, Hong Chang, MA Bingpeng, Shiguang Shan, and Xilin Chen.
\newblock Cross attention network for few-shot classification.
\newblock In {\em Advances in Neural Information Processing Systems (NeurIPS)},
  2019.

\bibitem{se}
Jie Hu, Li Shen, and Gang Sun.
\newblock Squeeze-and-excitation networks.
\newblock In {\em Proc. IEEE Conference on Computer Vision and Pattern
  Recognition (CVPR)}, 2018.

\bibitem{hu2018videomatch}
Yuan-Ting Hu, Jia-Bin Huang, and Alexander~G Schwing.
\newblock Videomatch: Matching based video object segmentation.
\newblock In {\em Proc. European Conference on Computer Vision (ECCV)}, 2018.

\bibitem{dccnet}
Shuaiyi Huang, Qiuyue Wang, Songyang Zhang, Shipeng Yan, and Xuming He.
\newblock Dynamic context correspondence network for semantic alignment.
\newblock In {\em Proc. IEEE International Conference on Computer Vision
  (ICCV)}, 2019.

\bibitem{batchnorm}
Sergey Ioffe and Christian Szegedy.
\newblock Batch normalization: Accelerating deep network training by reducing
  internal covariate shift.
\newblock In {\em Proc. International Conference on Machine Learning (ICML)},
  2015.

\bibitem{junejo2008cross}
Imran~N Junejo, Emilie Dexter, Ivan Laptev, and Patrick P{\'e}rez.
\newblock Cross-view action recognition from temporal self-similarities.
\newblock In {\em Proc. European Conference on Computer Vision (ECCV)}, 2008.

\bibitem{junejo2011view}
Imran~N Junejo, Emilie Dexter, Ivan Laptev, and Patrick P{\'e}rez.
\newblock View-independent action recognition from temporal self-similarities.
\newblock {\em IEEE Transactions on Pattern Analysis and Machine Intelligence
  (TPAMI)}, 33(1):172--185, 2011.

\bibitem{fcss}
Seungryong Kim, Dongbo Min, Bumsub Ham, Sangryul Jeon, Stephen Lin, and
  Kwanghoon Sohn.
\newblock Fcss: Fully convolutional self-similarity for dense semantic
  correspondence.
\newblock In {\em Proc. IEEE Conference on Computer Vision and Pattern
  Recognition (CVPR)}, 2017.

\bibitem{koch2015siamese}
Gregory Koch, Richard Zemel, and Ruslan Salakhutdinov.
\newblock Siamese neural networks for one-shot image recognition.
\newblock In {\em International Conference on Machine Learning (ICML) Workshop
  on Deep Learning}, 2015.

\bibitem{cifar}
Alex Krizhevsky and Geoffrey Hinton.
\newblock Learning multiple layers of features from tiny images.
\newblock Technical report, 2009.

\bibitem{motionsqueeze}
Heeseung Kwon, Manjin Kim, Suha Kwak, and Minsu Cho.
\newblock Motionsqueeze: Neural motion feature learning for video
  understanding.
\newblock In {\em Proc. European Conference on Computer Vision (ECCV)}, 2020.

\bibitem{stss}
Heeseung Kwon, Manjin Kim, Suha Kwak, and Minsu Cho.
\newblock Learning self-similarity in space and time as generalized motion for
  video action recognition.
\newblock In {\em Proc. IEEE International Conference on Computer Vision
  (ICCV)}, 2021.

\bibitem{metaoptnet}
Kwonjoon Lee, Subhransu Maji, Avinash Ravichandran, and Stefano Soatto.
\newblock Meta-learning with differentiable convex optimization.
\newblock In {\em Proc. IEEE Conference on Computer Vision and Pattern
  Recognition (CVPR)}, 2019.

\bibitem{ctm}
Hongyang Li, David Eigen, Samuel Dodge, Matthew Zeiler, and Xiaogang Wang.
\newblock Finding task-relevant features for few-shot learning by category
  traversal.
\newblock In {\em Proc. IEEE Conference on Computer Vision and Pattern
  Recognition (CVPR)}, 2019.

\bibitem{ancnet}
Shuda Li, Kai Han, Theo~W Costain, Henry Howard-Jenkins, and Victor Prisacariu.
\newblock Correspondence networks with adaptive neighbourhood consensus.
\newblock In {\em Proc. IEEE Conference on Computer Vision and Pattern
  Recognition (CVPR)}, 2020.

\bibitem{li2019revisiting}
Wenbin Li, Lei Wang, Jinglin Xu, Jing Huo, Yang Gao, and Jiebo Luo.
\newblock Revisiting local descriptor based image-to-class measure for few-shot
  learning.
\newblock In {\em Proc. IEEE Conference on Computer Vision and Pattern
  Recognition (CVPR)}, 2019.

\bibitem{lifchitz2021local}
Yann Lifchitz, Yannis Avrithis, and Sylvaine Picard.
\newblock Local propagation for few-shot learning.
\newblock In {\em International Conference on Pattern Recognition (ICPR)},
  2021.

\bibitem{dc}
Yann Lifchitz, Yannis Avrithis, Sylvaine Picard, and Andrei Bursuc.
\newblock Dense classification and implanting for few-shot learning.
\newblock In {\em Proc. IEEE Conference on Computer Vision and Pattern
  Recognition (CVPR)}, 2019.

\bibitem{negmargin}
Bin Liu, Yue Cao, Yutong Lin, Qi Li, Zheng Zhang, Mingsheng Long, and Han Hu.
\newblock Negative margin matters: Understanding margin in few-shot
  classification.
\newblock In {\em Proc. European Conference on Computer Vision (ECCV)}, 2020.

\bibitem{tpn}
Yanbin Liu, Juho Lee, Minseop Park, Saehoon Kim, Eunho Yang, Sung~Ju Hwang, and
  Yi Yang.
\newblock Learning to propagate labels: Transductive propagation network for
  few-shot learning.
\newblock In {\em Proc. International Conference on Learning Representations
  (ICLR)}, 2018.

\bibitem{luo2016efficient}
Wenjie Luo, Alexander~G Schwing, and Raquel Urtasun.
\newblock Efficient deep learning for stereo matching.
\newblock In {\em Proc. IEEE Conference on Computer Vision and Pattern
  Recognition (CVPR)}, 2016.

\bibitem{s2m2}
Puneet Mangla, Nupur Kumari, Abhishek Sinha, Mayank Singh, Balaji
  Krishnamurthy, and Vineeth~N Balasubramanian.
\newblock Charting the right manifold: Manifold mixup for few-shot learning.
\newblock In {\em IEEE Winter Conference on Applications of Computer Vision
  (WACV)}, 2020.

\bibitem{chm}
Juhong Min and Minsu Cho.
\newblock Convolutional hough matching networks.
\newblock In {\em Proc. IEEE Conference on Computer Vision and Pattern
  Recognition (CVPR)}, 2021.

\bibitem{hsnet}
Juhong Min, Dahyun Kang, and Minsu Cho.
\newblock Hypercorrelation squeeze for few-shot segmentation.
\newblock In {\em Proc. IEEE International Conference on Computer Vision
  (ICCV)}, 2021.

\bibitem{hpf}
Juhong Min, Jongmin Lee, Jean Ponce, and Minsu Cho.
\newblock Hyperpixel flow: Semantic correspondence with multi-layer neural
  features.
\newblock In {\em Proc. IEEE International Conference on Computer Vision
  (ICCV)}, 2019.

\bibitem{dhpf}
Juhong Min, Jongmin Lee, Jean Ponce, and Minsu Cho.
\newblock Learning to compose hypercolumns for visual correspondence.
\newblock In {\em Proc. European Conference on Computer Vision (ECCV)}, 2020.

\bibitem{relu}
Vinod Nair and Geoffrey~E Hinton.
\newblock Rectified linear units improve restricted boltzmann machines.
\newblock In {\em International Conference on Machine Learning (ICML)}, 2010.

\bibitem{oh2019video}
Seoung~Wug Oh, Joon-Young Lee, Ning Xu, and Seon~Joo Kim.
\newblock Video object segmentation using space-time memory networks.
\newblock In {\em Proceedings of the IEEE/CVF International Conference on
  Computer Vision}, pages 9226--9235, 2019.

\bibitem{oquab2014}
Maxime Oquab, Leon Bottou, Ivan Laptev, and Josef Sivic.
\newblock Learning and transferring mid-level image representations using
  convolutional neural networks.
\newblock In {\em Proc. IEEE Conference on Computer Vision and Pattern
  Recognition (CVPR)}, 2014.

\bibitem{tadam}
Boris Oreshkin, Pau~Rodr{\'\i}guez L{\'o}pez, and Alexandre Lacoste.
\newblock Tadam: Task dependent adaptive metric for improved few-shot learning.
\newblock In {\em Advances in Neural Information Processing Systems (NeurIPS)},
  2018.

\bibitem{pytorch}
Adam Paszke, Sam Gross, Soumith Chintala, Gregory Chanan, Edward Yang, Zachary
  DeVito, Zeming Lin, Alban Desmaison, Luca Antiga, and Adam Lerer.
\newblock Automatic differentiation in pytorch.
\newblock In {\em Advances in Neural Information Processing Systems (NeurIPS)
  Workshop Autodiff}, 2017.

\bibitem{qi2018low}
Hang Qi, Matthew Brown, and David~G Lowe.
\newblock Low-shot learning with imprinted weights.
\newblock In {\em Proc. IEEE Conference on Computer Vision and Pattern
  Recognition (CVPR)}, 2018.

\bibitem{tewam}
Limeng Qiao, Yemin Shi, Jia Li, Yaowei Wang, Tiejun Huang, and Yonghong Tian.
\newblock Transductive episodic-wise adaptive metric for few-shot learning.
\newblock In {\em Proc. IEEE International Conference on Computer Vision
  (ICCV)}, 2019.

\bibitem{qiao2018few}
Siyuan Qiao, Chenxi Liu, Wei Shen, and Alan~L Yuille.
\newblock Few-shot image recognition by predicting parameters from activations.
\newblock In {\em Proc. IEEE Conference on Computer Vision and Pattern
  Recognition (CVPR)}, 2018.

\bibitem{lsa}
Prajit Ramachandran, Niki Parmar, Ashish Vaswani, Irwan Bello, Anselm Levskaya,
  and Jon Shlens.
\newblock Stand-alone self-attention in vision models.
\newblock In {\em Advances in Neural Information Processing Systems (NeurIPS)},
  2019.

\bibitem{ravi2016optimization}
Sachin Ravi and Hugo Larochelle.
\newblock Optimization as a model for few-shot learning.
\newblock In {\em Proc. International Conference on Learning Representations
  (ICLR)}, 2017.

\bibitem{shotfree}
Avinash Ravichandran, Rahul Bhotika, and Stefano Soatto.
\newblock Few-shot learning with embedded class models and shot-free meta
  training.
\newblock In {\em Proc. IEEE International Conference on Computer Vision
  (ICCV)}, 2019.

\bibitem{tieredimagenet}
Mengye Ren, Eleni Triantafillou, Sachin Ravi, Jake Snell, Kevin Swersky,
  Joshua~B Tenenbaum, Hugo Larochelle, and Richard~S Zemel.
\newblock Meta-learning for semi-supervised few-shot classification.
\newblock In {\em Proc. International Conference on Learning Representations
  (ICLR)}, 2018.

\bibitem{ncnet}
Ignacio Rocco, Mircea Cimpoi, Relja Arandjelovi{\'c}, Akihiko Torii, Tomas
  Pajdla, and Josef Sivic.
\newblock Neighbourhood consensus networks.
\newblock In {\em Advances in Neural Information Processing Systems (NeurIPS)},
  2018.

\bibitem{rodriguez2020embedding}
Pau Rodr{\'\i}guez, Issam Laradji, Alexandre Drouin, and Alexandre Lacoste.
\newblock Embedding propagation: Smoother manifold for few-shot classification.
\newblock In {\em Proc. European Conference on Computer Vision (ECCV)}, 2020.

\bibitem{russakovsky2015imagenet}
Olga Russakovsky, Jia Deng, Hao Su, Jonathan Krause, Sanjeev Satheesh, Sean Ma,
  Zhiheng Huang, Andrej Karpathy, Aditya Khosla, Michael Bernstein, et~al.
\newblock Imagenet large scale visual recognition challenge.
\newblock {\em International Journal of Computer Vision (IJCV)},
  115(3):211--252, 2015.

\bibitem{leo}
Andrei~A Rusu, Dushyant Rao, Jakub Sygnowski, Oriol Vinyals, Razvan Pascanu,
  Simon Osindero, and Raia Hadsell.
\newblock Meta-learning with latent embedding optimization.
\newblock In {\em Proc. International Conference on Learning Representations
  (ICLR)}, 2018.

\bibitem{superglue}
Paul-Edouard Sarlin, Daniel DeTone, Tomasz Malisiewicz, and Andrew Rabinovich.
\newblock Superglue: Learning feature matching with graph neural networks.
\newblock In {\em Proc. IEEE Conference on Computer Vision and Pattern
  Recognition (CVPR)}, 2020.

\bibitem{schmidhuber1987evolutionary}
J{\"u}rgen Schmidhuber.
\newblock {\em Evolutionary principles in self-referential learning, or on
  learning how to learn: the meta-meta-... hook}.
\newblock PhD thesis, Technische Universit{\"a}t M{\"u}nchen, 1987.

\bibitem{facenet}
Florian Schroff, Dmitry Kalenichenko, and James Philbin.
\newblock Facenet: A unified embedding for face recognition and clustering.
\newblock In {\em Proc. IEEE Conference on Computer Vision and Pattern
  Recognition (CVPR)}, 2015.

\bibitem{shechtman2007matching}
Eli Shechtman and Michal Irani.
\newblock Matching local self-similarities across images and videos.
\newblock In {\em Proc. IEEE Conference on Computer Vision and Pattern
  Recognition (CVPR)}, 2007.

\bibitem{protonet}
Jake Snell, Kevin Swersky, and Richard Zemel.
\newblock Prototypical networks for few-shot learning.
\newblock In {\em Advances in Neural Information Processing Systems (NeurIPS)},
  2017.

\bibitem{sun2018pwc}
Deqing Sun, Xiaodong Yang, Ming-Yu Liu, and Jan Kautz.
\newblock Pwc-net: Cnns for optical flow using pyramid, warping, and cost
  volume.
\newblock In {\em Proc. IEEE Conference on Computer Vision and Pattern
  Recognition (CVPR)}, 2018.

\bibitem{sun2020mining}
Guolei Sun, Wenguan Wang, Jifeng Dai, and Luc Van~Gool.
\newblock Mining cross-image semantics for weakly supervised semantic
  segmentation.
\newblock In {\em Proc. European Conference on Computer Vision (ECCV)}, 2020.

\bibitem{mtl}
Qianru Sun, Yaoyao Liu, Tat-Seng Chua, and Bernt Schiele.
\newblock Meta-transfer learning for few-shot learning.
\newblock In {\em Proc. IEEE Conference on Computer Vision and Pattern
  Recognition (CVPR)}, 2019.

\bibitem{relationnet}
Flood Sung, Yongxin Yang, Li Zhang, Tao Xiang, Philip~HS Torr, and Timothy~M
  Hospedales.
\newblock Learning to compare: Relation network for few-shot learning.
\newblock In {\em Proc. IEEE Conference on Computer Vision and Pattern
  Recognition (CVPR)}, 2018.

\bibitem{inceptions}
Christian Szegedy, Wei Liu, Yangqing Jia, Pierre Sermanet, Scott Reed, Dragomir
  Anguelov, Dumitru Erhan, Vincent Vanhoucke, and Andrew Rabinovich.
\newblock Going deeper with convolutions.
\newblock In {\em Proc. IEEE Conference on Computer Vision and Pattern
  Recognition (CVPR)}, 2015.

\bibitem{tenenbaum1998mapping}
Joshua~B Tenenbaum.
\newblock Mapping a manifold of perceptual observations.
\newblock {\em Advances in Neural Information Processing Systems (NeurIPS)},
  1998.

\bibitem{rfs}
Yonglong Tian, Yue Wang, Dilip Krishnan, Joshua~B Tenenbaum, and Phillip Isola.
\newblock Rethinking few-shot image classification: a good embedding is all you
  need?
\newblock In {\em Proc. European Conference on Computer Vision (ECCV)}, 2020.

\bibitem{torabi2013local}
Atousa Torabi and Guillaume-Alexandre Bilodeau.
\newblock Local self-similarity-based registration of human rois in pairs of
  stereo thermal-visible videos.
\newblock {\em Pattern Recognition}, 46(2):578--589, 2013.

\bibitem{matchingnet}
Oriol Vinyals, Charles Blundell, Timothy Lillicrap, and Daan Wierstra.
\newblock Matching networks for one shot learning.
\newblock In {\em Advances in Neural Information Processing Systems (NeurIPS)},
  2016.

\bibitem{cub}
Catherine Wah, Steve Branson, Peter Welinder, Pietro Perona, and Serge
  Belongie.
\newblock The caltech-ucsd birds-200-2011 dataset.
\newblock Technical report, 2011.

\bibitem{wang2020video}
Heng Wang, Du Tran, Lorenzo Torresani, and Matt Feiszli.
\newblock Video modeling with correlation networks.
\newblock In {\em Proc. IEEE Conference on Computer Vision and Pattern
  Recognition (CVPR)}, 2020.

\bibitem{nlsa}
Xiaolong Wang, Ross Girshick, Abhinav Gupta, and Kaiming He.
\newblock Non-local neural networks.
\newblock In {\em Proc. IEEE Conference on Computer Vision and Pattern
  Recognition (CVPR)}, 2018.

\bibitem{wang2020few}
Xin Wang, Thomas~E. Huang, Trevor Darrell, Joseph~E Gonzalez, and Fisher Yu.
\newblock Frustratingly simple few-shot object detection.
\newblock In {\em Proc. International Conference on Machine Learning (ICML)},
  2020.

\bibitem{simpleshot}
Yan Wang, Wei-Lun Chao, Kilian~Q Weinberger, and Laurens van~der Maaten.
\newblock Simpleshot: Revisiting nearest-neighbor classification for few-shot
  learning.
\newblock {\em arXiv preprint arXiv:1911.04623}, 2019.

\bibitem{frn}
Davis Wertheimer, Luming Tang, and Bharath Hariharan.
\newblock Few-shot classification with feature map reconstruction networks.
\newblock In {\em Proc. IEEE Conference on Computer Vision and Pattern
  Recognition (CVPR)}, 2021.

\bibitem{vcn}
Gengshan Yang and Deva Ramanan.
\newblock Volumetric correspondence networks for optical flow.
\newblock {\em Advances in Neural Information Processing Systems (NeurIPS)},
  2019.

\bibitem{feat}
Han-Jia Ye, Hexiang Hu, De-Chuan Zhan, and Fei Sha.
\newblock Few-shot learning via embedding adaptation with set-to-set functions.
\newblock In {\em Proc. IEEE Conference on Computer Vision and Pattern
  Recognition (CVPR)}, 2020.

\bibitem{bmaml}
Jaesik Yoon, Taesup Kim, Ousmane Dia, Sungwoong Kim, Yoshua Bengio, and Sungjin
  Ahn.
\newblock Bayesian model-agnostic meta-learning.
\newblock In {\em Advances in Neural Information Processing Systems (NeurIPS)},
  2018.

\bibitem{yosinski2014transferable}
Jason Yosinski, Jeff Clune, Yoshua Bengio, and Hod Lipson.
\newblock How transferable are features in deep neural networks?
\newblock In {\em Advances in Neural Information Processing Systems (NeurIPS)},
  2014.

\bibitem{zbontar2016stereo}
Jure {\v Z}bontar and Yann LeCun.
\newblock Stereo matching by training a convolutional neural network to compare
  image patches.
\newblock {\em Journal of Machine Learning Research (JMLR)}, 17(1):2287--2318,
  2016.

\bibitem{deepemd}
Chi Zhang, Yujun Cai, Guosheng Lin, and Chunhua Shen.
\newblock Deepemd: Few-shot image classification with differentiable earth
  mover's distance and structured classifiers.
\newblock In {\em Proc. IEEE Conference on Computer Vision and Pattern
  Recognition (CVPR)}, 2020.

\bibitem{zhao2020exploring}
Hengshuang Zhao, Jiaya Jia, and Vladlen Koltun.
\newblock Exploring self-attention for image recognition.
\newblock In {\em Proc. IEEE Conference on Computer Vision and Pattern
  Recognition (CVPR)}, 2020.

\bibitem{zheng2021spatially}
Chuanxia Zheng, Tat-Jen Cham, and Jianfei Cai.
\newblock The spatially-correlative loss for various image translation tasks.
\newblock In {\em Proc. IEEE Conference on Computer Vision and Pattern
  Recognition (CVPR)}, 2021.

\bibitem{ziko2020laplacian}
Imtiaz Ziko, Jose Dolz, Eric Granger, and Ismail~Ben Ayed.
\newblock Laplacian regularized few-shot learning.
\newblock In {\em Proc. International Conference on Machine Learning (ICML)},
  2020.

\end{thebibliography}
